\documentclass[10pt,twocolumn,letterpaper]{article}

\usepackage{cvpr}              % To produce the CAMERA-READY version

\usepackage{multirow}
\usepackage[table]{xcolor}
\usepackage{amssymb} 
\usepackage{pifont}
\usepackage{tabularx}
\usepackage{amssymb}
\usepackage{adjustbox}
\usepackage{tabularx}

\usepackage{makecell} 
\usepackage{tabularx}
\usepackage{multirow}
\usepackage{caption}
\usepackage{float}
\usepackage{wrapfig}
\usepackage[table]{xcolor}

\definecolor{cvprblue}{rgb}{0.21,0.49,0.74}
\usepackage[pagebackref,breaklinks,colorlinks,allcolors=cvprblue]{hyperref}

\def\paperID{34002} % *** Enter the Paper ID here
\def\confName{CVPR}
\def\confYear{2026}

\title{Looking Beyond the Window: Global-Local Aligned CLIP \\ for Training-free Open-Vocabulary Semantic Segmentation}

\author{
ByeongCheol Lee, 
Hyun Seok Seong,
Sangeek Hyun,
Gilhan Park, 
WonJun Moon, 
Jae-Pil Heo\thanks{Corresponding author.} \\
Sungkyunkwan University \\
{\tt\small \{bc7817, gustjrdl95, hsi1032, a01152a, wjun0830, jaepilheo\}@skku.edu}
}

\begin{document}
\maketitle

\begin{abstract}
% 원본 {A sliding-window inference strategy is commonly adopted in recent training-free open-vocabulary semantic segmentation methods to overcome CLIP's limitation in processing high-resolution images.}
A sliding-window inference strategy is commonly adopted in recent training-free open-vocabulary semantic segmentation methods to overcome limitation of the CLIP in processing high-resolution images.
However, this approach introduces a new challenge: each window is processed independently, leading to semantic discrepancy across windows.
To address this issue, we propose Global-Local Aligned CLIP~(GLA-CLIP), a framework that facilitates comprehensive information exchange across windows. 
Rather than limiting attention to tokens within individual windows, GLA-CLIP extends key-value tokens to incorporate contextual cues from all windows.
% Nevertheless, we observe a window bias: outer-window tokens are less likely to be attended, since anchor features are produced through interactions within the inner window patches, thereby lacking semantic grounding beyond their local context
Nevertheless, we observe a window bias: outer-window tokens are less likely to be attended, since query features are produced through interactions within the inner window patches, thereby lacking semantic grounding beyond their local context.
% \SE{} % anchor feature --> query feature?
% To mitigate this, we introduce a proxy anchor, constructed by aggregating high-confidence tokens from all windows, which provides a unified semantic reference for measuring similarity across both inner- and outer-window patches.
To mitigate this, we introduce a proxy anchor, constructed by aggregating tokens highly similar to the given query from all windows, which provides a unified semantic reference for measuring similarity across both inner- and outer-window patches.
%원본: Furthermore, we propose a dynamic normalization scheme that adjusts attention strength according to object scale, mitigating the influence of irrelevant tokens, particularly in small-object scenarios. This is achieved by dynamically scaling and thresholding the attention map based on the distribution of similarity scores.
Furthermore, we propose a dynamic normalization scheme that adjusts attention strength according to object scale by dynamically scaling and thresholding the attention map to cope with small-object scenarios.
Moreover, GLA-CLIP can be equipped on existing methods and broad their receptive field. Extensive experiments validate the effectiveness of GLA-CLIP in enhancing training-free open-vocabulary semantic segmentation performance. Codes are available at \href{https://github.com/2btlFe/GLA-CLIP}{github.com/2btlFe/GLA-CLIP}

\end{abstract}

% In the current baseline, ProxyCLIP exhibits a high Boundary Error Rate (BER), which quantifies the proportion of pixel pairs that are identical in the ground truth but inferred differently across window boundaries. As shown in (a), ProxyCLIP suffers from significant inconsistencies near window edges, leading to elevated BER values. In contrast, our proposed GLA-CLIP facilitates effective information exchange across windows, substantially reducing the Boundary Error Rate across all datasets. Furthermore, as illustrated in (b), GLA-CLIP suppresses unnatural grid artifacts commonly observed in ProxyCLIP, resulting in smoother and more coherent segmentation outputs.

\begin{figure}[t]
\centering
\vspace{-0.2cm}
\includegraphics[width=0.9\columnwidth]{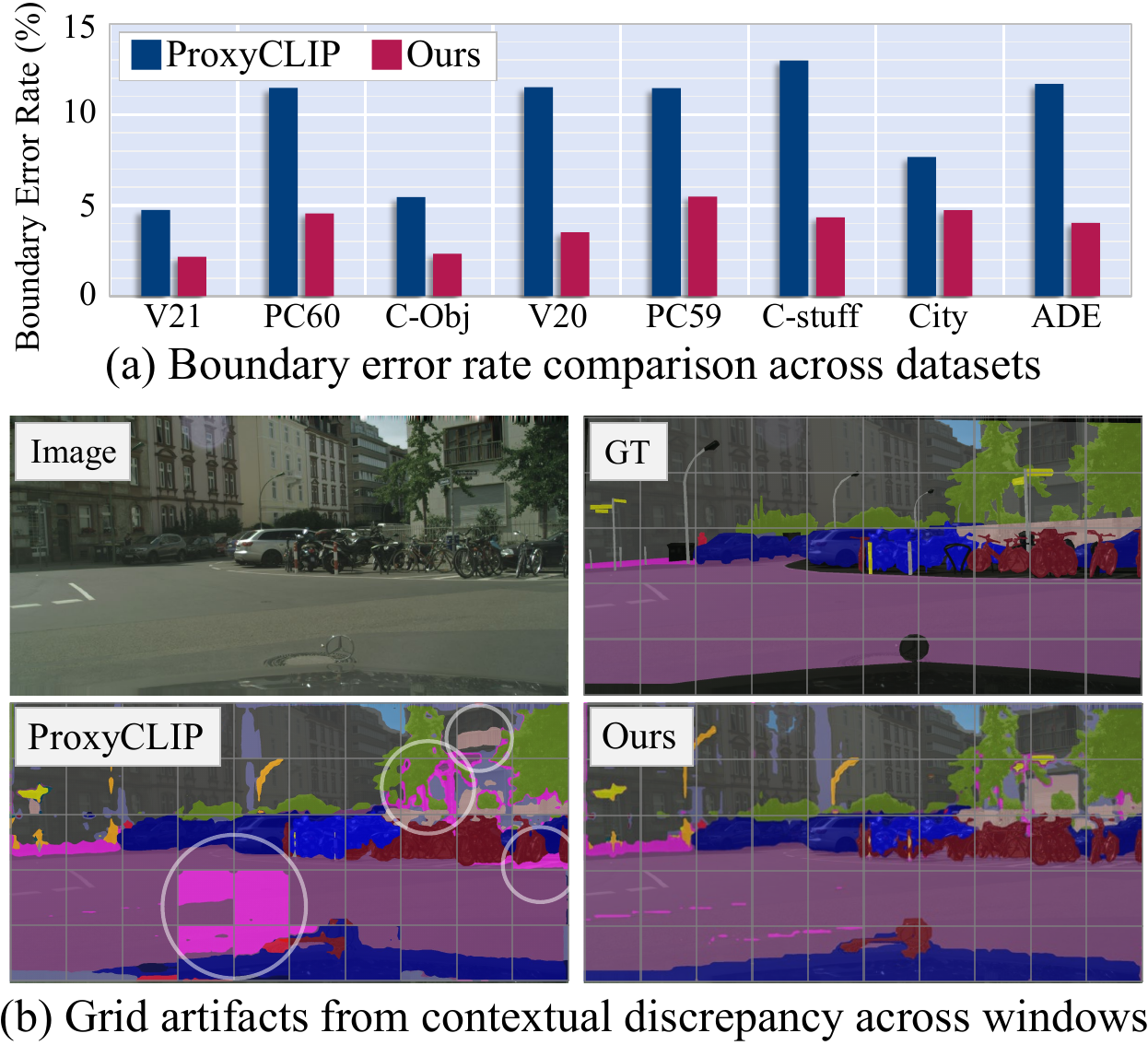} 
\vspace{-0.1cm}
\caption{ 
% (a) ProxyCLIP shows a high Boundary Error Rate (BER), measuring how often identical GT pixels are inferred differently across window boundaries. In contrast, our GLA-CLIP encourages cross-window information exchange, significantly reducing BER across all datasets. (b) GLA-CLIP effectively suppressing grid artifacts in ProxyCLIP. The yellow box indicates the grid arifacts where there is no grid artifacts in our model.
% Comparison of cross-window consistency and visual artifacts.
% (a) We measure Boundary Error Rate~(BER) across eight datasets to quantify the inconsistency of predictions along overlapping window boundaries. BER is computed as error rate on shared boundary pixels with identical ground-truth labels. ProxyCLIP exhibits a significantly higher BER due to its lack of inter-window reasoning, while our method, GLA-CLIP, achieves substantially lower BER by enabling effective cross-window attention.
% (b) Visual comparison of grid artifacts. ProxyCLIP suffers from severe grid-like errors~(highlighted in white circles), caused by insufficient context within individual windows. In contrast, GLA-CLIP suppresses these artifacts by leveraging outer-window contextual cues. 
Comparison of segmentation consistency near window boundaries.
% and visual artifacts.
(a) We evaluate segmentation inconsistency using the Boundary Error Rate~(BER).
% across eight datasets. 
BER is defined as the proportion of pixels near adjacent window boundaries where predicted labels differ despite identical ground-truth.
% (a) We evaluate the rate of segmentation inconsistency by measuring the Boundary Error Rate~(BER) across eight datasets. BER is defined as the proportion of pixels located near the boundaries between adjacent windows where the predicted labels differ, despite having identical ground-truth annotations. 
% A high BER indicates inconsistent predictions along window boundaries.
ProxyCLIP yields high BER due to its lack of cross-window interaction, whereas BER is significantly reduced in ours by incorporating global context into attention process. More details in Appendix~\ref{sec:BER}.
(b) ProxyCLIP exhibits grid artifacts~(marked with white circles), caused by the limited receptive field within individual windows. 
In contrast, ours mitigates these artifacts by leveraging contextual information beyond local windows. 
% \textcolor{blue}{본문에 fig1 언급할 곳 찾아서 ref달기}
}
\label{fig: motivation BER}
\vspace{-0.3cm}
\end{figure}
\section{Introduction}

Open-vocabulary semantic segmentation~(OVSS) aims to assign pixel-level semantic labels from an open and potentially unbounded vocabulary, enabling models to generalize beyond a fixed set of training categories. 
Recent advances in vision-language models, such as CLIP~\cite{CLIP}, have made it possible to approach this task without additional training, by leveraging the image-text aligned embedding space. 
This has led to a growing interest in Training-Free OVSS~\cite{maskclip, gem, sclip, clearclip, naclip, lan2024proxyclip, cliptrase, clipseg, resclip}, where segmentation is performed only with minimal modifications to the CLIP architecture or processing pipeline.

%원본: However, a fundamental limitation remains: CLIP is pretrained on low-resolution inputs~(e.g., 224×224), which hinders its applicability to high-resolution segmentation tasks.

However, a fundamental limitation remains: CLIP is pretrained on low-resolution inputs~(e.g., 224×224), which hinders its applicability to high-resolution segmentation tasks.
To address this, recent works adopt a sliding-window strategy~\cite{maskclip, sclip, clearclip, naclip, lan2024proxyclip, CASS}, where high-resolution images are divided into overlapping crops and processed independently.
By extending precise text-aligned embeddings to high-resolution settings without compromising spatial details, this simple approach has led to a significant performance boost.
%원본: However, we argue that this strategy still has a critical limitation: each window lacks access to the broader scene context, which leads to inconsistent predictions across windows.
However, we argue that this strategy still has a critical limitation: each window lacks access to the broader scene context, which leads to inconsistent predictions across windows.
% While this approach enhances CLIP in handling high-resolution images, this approach introduces new challenges: each window lacks access to the broader scene context, leading to fragmented predictions, inconsistent labeling across windows, and severe artifacts at object boundaries.

% Fig. 1 언급
One clear example of this limitation is illustrated in Fig.~\ref{fig: motivation BER}, where independently processed windows produce inconsistent predictions near window boundaries and introduce grid-like artifacts. Specifically, adjacent windows often assign different semantic labels to neighboring pixels along shared boundaries, even when those pixels belong to the same ground-truth class, resulting in visible discontinuities and structured artifacts. These issues highlight the absence of global context integration and motivate the need for a mechanism that enforces semantic consistency across windows.

To address this issue, we propose Global-Local Aligned CLIP~(GLA-CLIP), a novel training-free framework that explicitly aligns local and global semantics.
We begin by introducing a Key-Value Extension mechanism that allows each query to attend to an unbounded set of key-value tokens across the entire image.
This design enables broader context integration, thereby mitigating the semantic discrepancy across independently processed windows.
However, we observe that even with globally extended key-value tokens, the attention remains locally biased: query tokens disproportionately attend to patches within the same window while underweighting semantically similar tokens outside the window.
This is attributed to the fact that query embeddings are constructed solely from inner-window features, limiting their global awareness.
% To alleviate this issue, we derive proxy tokens to serve as globally representative queries by aggregating the high-confidence patches across the entire image.
To alleviate this issue, we derive proxy tokens to serve as globally representative queries by aggregating the patches highly similar to the given query across the entire image.
These proxies help reduce local bias and encourage semantically aligned attention across windows.
Finally, we introduce a dynamic attention normalization strategy that adjusts attention magnitudes based on object scale, estimated from each query's similarity distribution.
This adaptation ensures that small objects are not overwhelmed by irrelevant global tokens, while large objects benefit from richer contextual clues.
Our results across multiple benchmarks demonstrate the effectiveness of our proposed approaches in mitigating the semantic discrepancy between different sliding windows.

Our contributions are summarized as follows:
\begin{itemize}
\item To our knowledge, this is the first work to identify and address the challenge stemming from the sliding window: locally bounded semantics in each window incur inconsistent prediction between windows.
\item We extend the set of key-value tokens to cover the entire image, allowing each query to reason over global context regardless of its local window.
\item We introduce globally representative proxy queries by aggregating high confidence tokens, enabling semantically consistent attention across windows.
\item We propose a dynamic attention normalization mechanism that adapts the influence of global tokens based on the estimated object scale, improving robustness across varying object sizes and domains.
\item Our framework can expand the receptive field of any baseline model without additional training.

\end{itemize}

\begin{figure*}
    \centering
    \vspace{-0.2cm}
    \includegraphics[width=1.\textwidth]{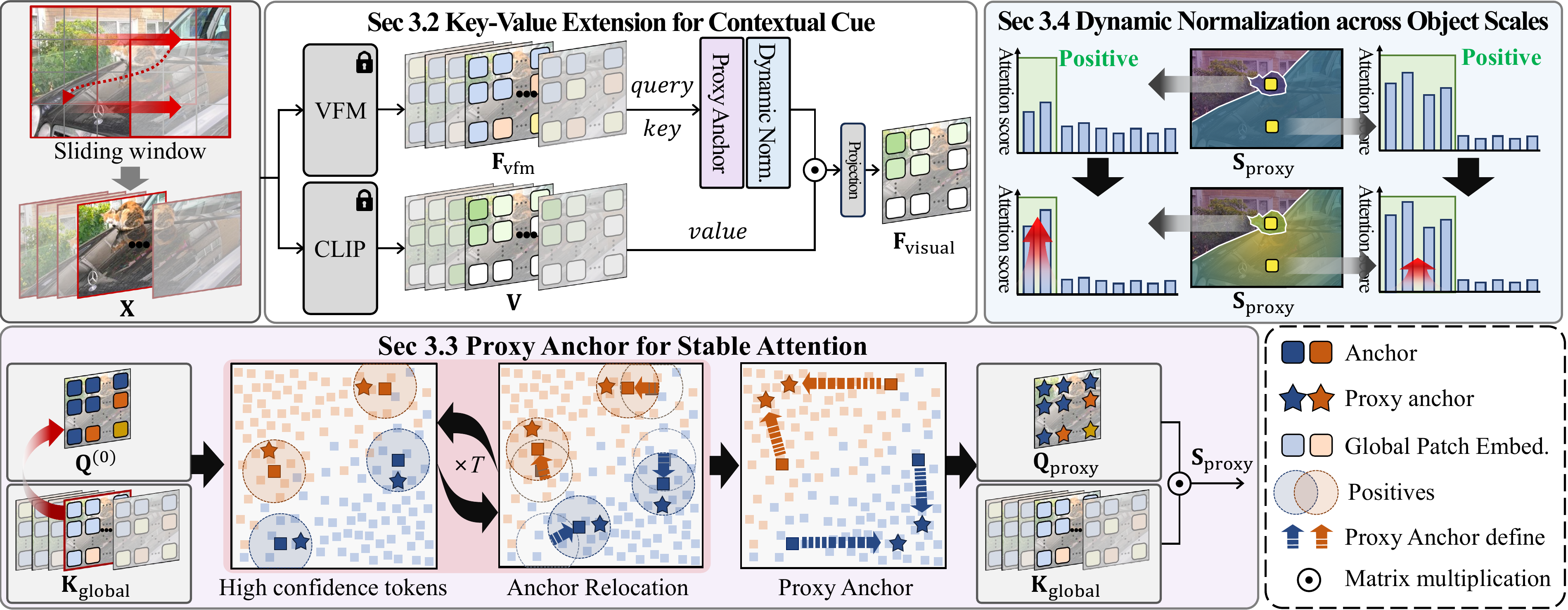}
    \vspace{-0.5cm}
    \caption{
        Overview of our proposed framework.
        The input image is first divided into overlapping windows and processed using frozen backbones: a Vision Foundation Model~(VFM, e.g., DINO) and CLIP. We introduce a Key-Value Token Extension, where VFM features from the current window $\mathbf{F}_{\text{vfm}}$ serve as \textit{query} tokens, while \textit{key} tokens are gathered from all windows to provide global context. The corresponding \textit{value} tokens $V$ are extracted from the final transformer layer of CLIP. Cross-attention is then applied, followed by a projection layer to generate the final visual features $\mathbf{F}_{\text{visual}}$. To stabilize attention across windows, each query token is replaced with a semantically representative proxy anchor. Finally, a dynamic normalization scheme adjusts attention strength based on object size, approximated by the number of positive samples associated with each proxy anchor.
        % \HS{}
        % 1. copy 라는 말좀 이상하다. K_global의 window 하나를 테두리 진하게 칠해서 화살표 땡겨서 위로 올리는 식으로 표현하면 될듯.
        % 2. propagation이라는 표현은 논문에 쓰지 않았어서 'anchor define' 이런식으로 쓰는게 나을듯
        % 3. positiveness criterion 용어도 본문과 맞추기 - 
        % 4. sec3.3 3.4를 q,k만 가지고 하는거니까 저 cross attention 네모박스 크기를 작게하고 위에꺼만 인풋되게 하면좋을듯 - 해결
        % 1. T는 기울기, 2. 동그라미 범위를 positiv라고 하면좋을듯, 3. projection layer
    }
    \label{fig:Overview}
    \vspace{-0.2cm}
\end{figure*}

\section{Related Work}
\subsection{Open-Vocabulary Semantic Segmentation}
The emergence of pre-trained vision-language models~\cite{CLIP, ALIGN} enables aligning text and image features for open-vocabulary semantic segmentation~\cite{clipseg2,DenseClip}. Early methods follow a two-stage pipeline, where class-agnostic masks are first generated and then matched with text embeddings using models like CLIP~\cite{Zegformer, maskclip_train, Zseg, Ovseg, SegCLIP}. While these methods are effective, they depend on external mask generators that are separate from the vision-language model. This separation can reduce visual-semantic consistency. To address this, single-stage methods~\cite{zegclip, FCCLIP, SAN} have been introduced. These methods directly predict segmentation masks from CLIP features and often use techniques such as attention bias~\cite{SAN} or mask pooling~\cite{FCCLIP}. Other approaches include ODISE~\cite{ODISE}, which uses frozen diffusion features with a Mask2Former decoder~\cite{mask2former}, and CAT-Seg~\cite{cat-seg}, which improves alignment by pixel-text similarity and text encoder fine-tuning. More recently, EBSeg~\cite{ebseg} enhances generalization by adjusting image-text similarity distribution during training. Meanwhile, training-free approaches have gained attention for utilizing CLIP without requiring further training.
\subsection{Training-Free OVSS}
Training-free open-vocabulary semantic segmentation makes predictions without any additional training or fine-tuning. Instead of introducing learnable parameters, these approaches enhance the segmentation ability of CLIP by modifying its image encoder~\cite{CLIPsurgery, wang2024sclip} and leveraging vision foundation models~\cite{diffusionSeg,lan2024proxyclip}. MaskCLIP~\cite{maskclip} enhances spatial sensitivity in CLIP by discarding the query and key embeddings in the final self-attention block, using only the value embeddings. Subsequent methods have modified the attention mechanism by replacing query-key attention with value-value or self-self attention~(e.g., query-query, key-key)~\cite{CLIPsurgery, gem, sclip, cliptrase}. In addition, some approaches remove the final feed-forward layer and residual connections, as these components have been found to introduce noise in the resulting segmentation masks~\cite{naclip, clearclip}. ProxyCLIP~\cite{lan2024proxyclip} introduces a proxy attention mechanism that enhances spatial and semantic alignment in CLIP. However, most methods rely on sliding-window inference due to low training resolution, which makes it difficult for the model to understand the entire image and often leads to inconsistent results across different windows. To address these challenges, our method enhances global context and consistency by introducing key-value token extension and proxy-similarity. In doing so, it effectively mitigates attention imbalance and overcomes the inherent limitations of sliding-window inference.

\section{Method}

\subsection{Preliminary}
% 1. CLIP's representation
% 2. CLIP ViT, Attention 수식
% 3. Locality를 부여하기 위해, 
% CLIP의 마지막 레이어의 Attention 자리에 Self-Similarity Map을 적용하고, FFN, Residual 을 삭제하여 Segmentation에 적용할 수 있게 되었다. 

% CLIP-ViT는 visual encoder와 text encoder로 구성되어 있다. 
% Visual encoder의 경우, Attention 연산 수행, q, k, v 생성, 기존에는 q, k로 attention을 만들었지만, segmentation task를 수행하기 위해, locality를 반영한 Attention을 강제로 넣어줘야 했다. 이 과정에서 VFM의 feautre self-similarity map을 활용했고 (ProxyCLIP), residual고 FFN은 삭제했다. 

% 여기에서 전체 이미지 (I)를  patchify해서 N개의 토큰으로 바꾼다는 내용
% Self Similarity를 표현하는 식 추가해야 한다.
% Self-Similarity Map of VFM이 Self-Similarity Map of Q, K, V 보다 더 좋았다는 내용을 넣을 수도 있다. (Object 단위로 더 잘 뭉치더라)

% -----------------------------------------------------------------

% 1. Dense inference
% 이미지를 CLIP의 visual encoder를 통과시켜서 CLIP visual feature를 얻고, 
% 주어진 category 텍스트들을 text encoder에 통과시켜서 CLIP text feature를 얻는다. 
% 이후 dense inference는 각 visual 토큰과 text 토큰으로 cossim을 구해서 logit을 얻은 뒤
% 가장 높은 class로 결정한다:

% 용어 통일:
% Visual token
% Text token

\subsubsection{Open-Vocabulary Semantic Segmentation~(OVSS)}
% Open-vocabulary semantic segmentation leverages the visual-language alignment of CLIP \cite{CLIP}. Given an image \(\mathbf{X}\), the CLIP visual encoder divides it into N patches and converts the patches into a visual tokens \(\mathbf{F}_{\text{visual}}\). \SE{0731/10AM} % patch 라고 할거면 모든 글에서 패치라고 하던가, window라고 하려면 window라고 하던가 통일성 있게 해주세요
OVSS leverages the visual-language alignment of CLIP \cite{CLIP}. Given an image $\mathbf{X}$, the CLIP visual encoder converts it as visual tokens \(\mathbf{F}_{\text{visual}} \in \mathbb{R}^{N \times D}\), where $N$ is the number of tokens and $D$ is dimension size.
Meanwhile, the text encoder processes a set of text prompts to project them to a text tokens \(\mathbf{F}_{\text{text}}\). The cosine similarity between visual and text tokens determines logits, eventually categorizing each visual token to the text corresponding to the highest logits. %similarity score.

% 더 줄인다.
% 1. OVSS CLIP을 사용한다.
% 2. 2가지 타입의 CLIP encoder를 이용해 이미지 X를 N개의 patch로 나눠서 visual token을 만들고 a set of text prompts를 text token으로 변환한다.
% 3. CLIP visual token과 text token의 cosine_similarity로 logit을 계산하고 가장 높은 similarity score의 카테고리를 해당 토큰의 카테고리로 결정한다.  

% \subsubsection{ProxyCLIP} \HS{}%Baseline 으로 이름바꾸는거어떰
% CLIP visual feature를 뽑되 dense inference가 가능하게 뽑으려고 ProxyCLIP 모델을 베이스라인으로 삼았다. 
% ProxyCLIP은 CLIP의 마지막 layer에서 기존의 attention map이 아니라 VFM의 feature, 특히 DINO의 feature를 뽑아서
% 해당 feature의 self-similarity map을 attention map으로 적용하였다. 구체적으로 그 과정에서 Normalization을 거쳐서 적용을 했다. shifting과 scaling factor를 추가해서 정했고 이후, masking을 통해 unrelevent token을 상당 수 제거했다. 구체적인 수식은 다음과 같다:

% 더 줄인다. 
% 1. CLIP visual token을 segmentation에 적합하게 추출하기 위해 Training-Free OVSS model이 있었고, 그 중 우리는 ProxyCLIP 모델을 베이스라인으로 삼았다. 
% 2. ProxyCLIP은 CLIP의 마지막 transformer block의 Attention Map을 VFM feature의 self-similarity map으로 대체하였다. DINO를 주로 사용했다. 
% 3. 그리고 Attention Map의 경우에는 dino의 feature의 self-similarity map을 이용했다. 이 과정에서 normalization과 masking을 진행하여 더 정밀한 Attention Map을 구축한 모델이다  
% 4. 최종적으로는 visual token을 생성하기 위해 attention map과 Value 간의 multiplication을 진행한 뒤, 마지막 transformer block의 Projection layer를 거쳐 visual token을 추출하게 된다.
Recently, ProxyCLIP~\cite{lan2024proxyclip} has shown notable advances in OVSS by leveraging Vision Foundation Model~(VFM) features.
% A notable ad
We adopt ProxyCLIP as our baseline, which replaces the attention map in the final transformer block of CLIP with a self-similarity map computed from features of a VFM. 
Specifically, the self-similarity map $\mathbf{S}$ is defined as:
\begin{align}
\mathbf{S} &= \mathbf{F}_{\text{vfm}} \cdot \mathbf{F}_{\text{vfm}}^{\top} \in \mathbb{R}^{N \times N},
\label{eq:self_similarity}
\end{align}
where $(\cdot)$ indicate matrix multiplication and $\mathbf{F}_\text{vfm}$ denotes the visual features extracted from a VFM~(e.g., DINO~\cite{DINO}), which serves as the \textit{query} and \textit{key} embeddings in the attention mechanism.
% We represent VFM features as \(\mathbf{F}_\text{vfm}\). We adopt DINO \cite{DINO} as our VFM, using its self‑similarity map to replace the original attention in the last transformer block of CLIP. Self-similarity map is represented as:

% \begin{align}
% \mathbf{S} &= \mathbf{F}_{\text{vfm}} \mathbf{F}_{\text{vfm}}^{\top} \in \mathbb{R}^{N \times N}
% \label{eq:self_similarity}
% \end{align}

% Additionaly, ProxyCLIP adds normalization and masking technique to modify the self-similarity map. Specifically, normalization and masking equations are arranged as:

% In addition, for robustly handling the distributional differences of self-similarity maps across various VFMs, additional normalization and masking are applied to their self-similarity maps as follows:
To better integrate the self-similarity map into CLIP for open-vocabulary segmentation, they modulate and threshold the similarity scores to integrate semantically positive tokens. This refinement suppresses irrelevant token interactions and enables the similarity map to function as an effective attention map for segmentation tasks:
\begin{align}
\mathbf{A} &= \gamma \left( \mathbf{S} - \frac{\beta}{N^2} \sum_{i,j} \left[ \mathbf{S} \right]_{ij} \right), \label{eq:norm} \\
\mathcal{M}_{ij} &= 
\begin{cases} 
0, & A_{ij} \ge 0 \\
-\infty, & A_{ij} < 0,
\end{cases} \label{eq:masking} \\
\mathbf{Attn} &= \operatorname{SoftMax}(\mathbf{A} + \mathcal{M}),
\label{eq:proxy_attention}
\end{align}
where $\beta$ and $\gamma$ are fixed shift and scale hyperparameters.

The refined attention map $\mathbf{Attn}$ is applied to the \textit{value} tokens $\mathbf{V} \in \mathbb{R}^{N \times D}$ from the final transformer block of CLIP. The final output is obtained via a projected matrix multiplication, following prior findings that skip residual and feed-forward layers~\cite{clearclip}:
\begin{align}
\mathbf{F}_{\text{visual}} = \text{Proj}( \mathbf{{Attn} }  \cdot \mathbf{V}),
\label{eq:visual_token}
\end{align}
where $\text{Proj}$ denotes the projection layer from the last transformer block of CLIP.

\begin{figure*}
    \centering
    \vspace{-0.2cm}
    \includegraphics[width=1\textwidth]{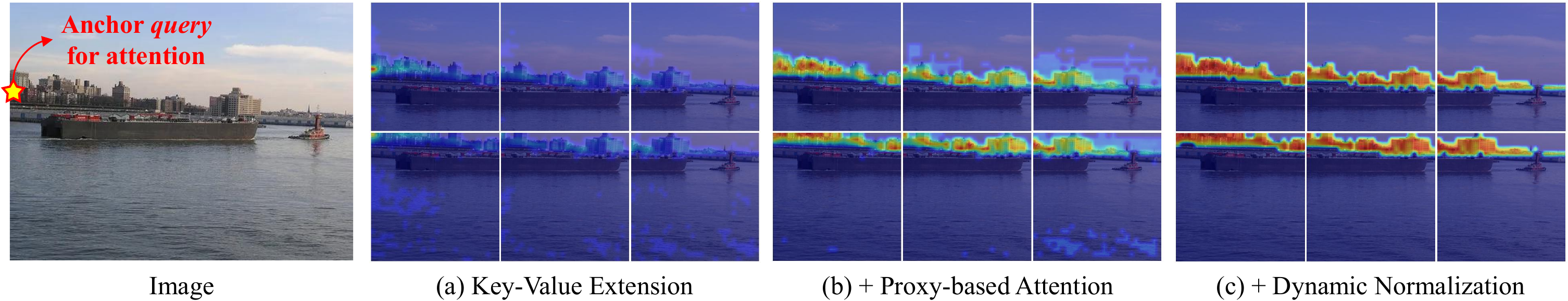}
    \vspace{-0.6cm}
    \caption{
        Visualization of attention maps for an anchor \textit{query} token. Proxy-based attention enhances focus on semantically relevant regions across both inner- and outer-window areas. The subsequent dynamic normalization further suppresses irrelevant responses, especially from noisy tokens, yielding sharper and more semantically consistent attention distributions.
    }
    \label{fig:attention_ablation_viz}
    \vspace{-0.2cm}
\end{figure*}

\subsubsection{Sliding-window}
% 3. Image preprocessing
% 일반적으로 사용하는 이미지의 경우 CLIP의 pre-trained resolution인 224 x 224 보다 큰 경우가 많아서 Detail한 부분까지도 inference를 잘 하기 위해 sliding-window를 적용한다. 이로 인해 우리가 다뤄야 할 이미지는 하나가 아니라 L개의 subwindows가 된다. 이후 각 window에 대해 독립적으로 ovs를 진행한 뒤 window끼리 중복되는 위치에 대해서는 각 logit을 모아 평균을 계산해서 최종 logit을 결정한다.  
% To effectively handle high-resolution images, we employ an overlapped sliding-window strategy that divides the original image into smaller sub-windows. Each window maintains pre-trained resolution of CLIP, which is 224$\times$224 pixels. Specifically, the original image is divided into $L$ sub-windows, 
% % and the aggregated sub-windows are represented as $\mathcal{X} = \{X_{1}, X_{2}, ..., X_{L}\}$. 
% and open-vocabulary semantic segmentation is carried out on every sub‑window independently. When multiple windows cover the same patch in original image, their logits are averaged, and the resulting mean is used as the final per‑patch prediction.
Since CLIP is pre-trained with a fixed input resolution of $224\times224$ pixels, it cannot directly process high-resolution inputs. Conventionally, this limitation is addressed by applying an overlapped sliding-window strategy, where the input image is divided into $L$ sub-windows of CLIP-compatible size. We follow this approach and perform OVSS independently on each sub-window. For overlapping regions, the logits are averaged across windows to produce the final per-patch predictions.

\subsection{Key-Value Extension for Contextual Cue}
\label{sec:key_value_extension}
While the sliding-window inference strategy is effective for adapting CLIP to high-resolution inputs by preserving its pre-trained resolution, it inherently restricts the model’s receptive field to local window regions. This constrained field of view prevents the model from accessing holistic scene-level context, thereby hindering its ability to aggregate semantically relevant cues across spatially distant regions. The limitation becomes especially pronounced when semantically coherent regions~(such as large objects or spatially continuous stuff categories) are partitioned across multiple windows. Such contextual discontinuity often leads to inconsistent or erroneous semantic predictions, as shown in Fig.~\ref{fig: motivation BER}.
% \SE{0801/10AM}
% 1. --> ref로 헷갈릴까봐 체크해놓기
% 위 주장이 맞나요? 윈도우 별로 잘못 inference되는게 문제 아닌가? entirely incorrect 란 말이 윈도우 내부에서 잘못된다는거면 명시해주면 좋을 듯 .
% 그리고 큰 물체의 일부만 봐서 문제가 생긴다 --> 이 말이 덜 강조되긴 하는 듯? limits the model to partial views of the image <- 이거만 쓰기에는 메인 주장인데 약한거 같음

% To address this limitation, we propose a key-value token extension strategy that augments the attention mechanism with information beyond individual windows. In traditional sliding-window approaches, each window is processed independently, preventing the model from leveraging cues outside the local region. This often results in fragmented or inconsistent predictions, especially for large objects or context-dependent categories.
% Our method explicitly injects global contextual information into each window’s attention operation by extending its receptive field to include tokens gathered from all windows. 
% Specifically, we collect the visual features extracted from the VFM, to serve as the global \textit{key} embeddings, and simultaneously collect the corresponding \textit{value} tokens from the final transformer layer of CLIP.
% By aggregating these tokens across all $L$ windows, we construct a unified set of global \textit{key} $\mathbf{K}_{\text{global}}$ and \textit{value} tokens $\mathbf{V}_{\text{global}}$ as follows:
To enhance contextual awareness in attention computation, we propose a key-value token extension, which expands the attention scope of each window by allowing it to reference visual information from the entire image, not just its local region.
Specifically, we collect the visual features extracted from the VFM to serve as global \textit{key} embeddings, and simultaneously retrieve the corresponding \textit{value} tokens from the final transformer layer of CLIP. By aggregating these tokens across all $L$ windows, we construct a unified set of global \textit{key} $\mathbf{K}_{\text{global}}$ and \textit{value} tokens $\mathbf{V}_{\text{global}}$ as follows:
\begin{align}
\mathbf{K}_{\text{global}} &= [\mathbf{F}_\text{vfm}^{(1)}; \mathbf{F}_\text{vfm}^{(2)}; \dots; \mathbf{F}_\text{vfm}^{(L)}] \in \mathbb{R}^{(L N) \times D}, \\
\mathbf{V}_{\text{global}} &= [\mathbf{V}^{(1)}; \mathbf{V}^{(2)}; \dots; \mathbf{V}^{(L)}] \in \mathbb{R}^{(LN) \times D}.
\end{align}
% These global tokens allow the query of each window to access a broader semantic context, enabling consistent and context-aware attention across the entire image.

% Subsequently, by performing conventional attention using the \textit{query} tokens $\mathbf{Q}$ from the current window and the extending \textit{key} and \textit{value} tokens ($\mathbf{K}_{\text{global}}, \mathbf{V}_{\text{global}}$), we obtain the context-aware visual tokens for the current window:
With these global tokens, we perform cross-window attention by computing similarity between the local \textit{query} tokens from the current window and the aggregated global \textit{key} tokens, followed by weighted aggregation over the corresponding \textit{value} tokens:
\begin{align}
\mathbf{A}_{\text{ext}} &= \mathbf{Q} \cdot \mathbf{K}_{\text{global}}^\top \in \mathbb{R}^{N \times (LN)}, \\
\mathbf{F}_{\text{visual}} &= \text{Proj}( \mathbf{A}_{\text{ext}}  \cdot \mathbf{V}_\text{global})\in \mathbb{R}^{N \times D}.
\label{eq:extended_visual_token}
\end{align}
Here, $\mathbf{Q} \in \mathbb{R}^{N \times D}$ denotes the set of query tokens extracted from the $\mathbf{F}_\text{vfm}$ of the current sliding window. These serve as the local anchors in the attention operation and initiate semantic integration with globally gathered keys and values.

\subsection{Proxy Anchor for Stable Attention}
\label{sec:proxy_anchor}
After applying the key-value token extension, we observe a persistent locality bias: attention is predominantly concentrated on tokens within the same window~(inner-window), while downweighting semantically similar tokens from other windows~(outer-window), as shown in Fig.~\ref{fig:attention_ablation_viz}(a).
% \HS{} % 증거 table이나 fig 만들어서 reference
Notably, even when outer-window tokens correspond to the same spatial location in the original image (i.e., overlapping regions), attention remains biased toward inner-window tokens. This behavior limits the model’s capacity to aggregate consistent semantics across windows.
To alleviate this bias, we introduce a proxy-based attention mechanism that replaces each query token with a semantically representative proxy aggregated from tokens with high similarity across all windows. This proxy serves as a stable anchor that reflects the underlying semantics of the region, allowing attention to be allocated based on semantic coherence rather than window-local bias. Consequently, our method promotes more globally consistent attention patterns and enhances cross-window integration.
% \SE{comment 0801/10AM} % 저한테는 목표가 불분명해보여요. imbalanced attention (혹은 locality bias)가 왜 proxy를 통해서 해결되는지가 단순히 semantic을 더 잘 반영하는 쿼리 토큰으로 변환해서다? <-- 이게 좀 이상하게 들림
% 예를 들어서, 뒷 단락에서 나오듯, unsup으로 할거면 애초에 dino feature 공간 상에서 Similarity를 잰다는건데 이게 왜 단순 attention에서 생기는 locality bias를 효과적으로 해결하냐? 에 대한 설명이 더 필요해보임
% 예를 들어, 어떤 repeatedly extended 하는 알고리즘의 특성이 locality를 해결한다던지 하는 명확한 이유?

% While various methods exist for computing such proxies~\cite{kim2020proxy, li2020prototypical, PPAP}, we follow a non-parametric, unsupervised strategy~\cite{PPAP}. 

Specifically, for a given query token, we follow \cite{PPAP} to repeatedly compute the mean vector of high-similarity neighbors, gradually converging on a proxy that best represents the shared semantics around the anchor.
We define the initial proxy-query \(\mathbf{Q}_{i}^{(0)}\), which stems from \(i\)-th token of local query tokens \(\mathbf{Q}\) and construct the index set of high-confident tokens \(\mathcal{P}_i^{(0)}\), which indicates a higher cosine similarity compared to a pre-defined threshold $\rho$ as below:
\begin{align}
% \mathbf{q}_i^{(0)} &= \mathbf{Q}_{\text{local}}[i] \in \mathbb{R}^{D} \label{eq:init_feat} \\
\mathcal{P}_i^{(0)} &= \left\{ j \,\middle|\, \mathbf{Q}_i^{(0)} \cdot \mathbf{K}_{\text{global},j} > \rho,\ j \in \mathcal{J} \right\}, \label{eq:positive_set}
\end{align}
where $\mathcal{J}$ denotes the index set comprising all patch features of all windows.
Also, the $t$-th proxy is defined as follows:
\begin{align}
\mathbf{Q}_i^{(t)} &= \frac{1}{\left| \mathcal{P}_i^{(t-1)} \right|} \sum_{j \in \mathcal{P}_i^{(t-1)}} \mathbf{K}_{\text{global},j}.
\label{eq:proxy}
\end{align}
By repeating $T$ steps of Eq.~\eqref{eq:positive_set} and \eqref{eq:proxy} for generating refined proxy, we obtain $\mathbf{Q}_i^{(T)}$.
This operation is computed for all \textit{query} tokens.
Using the proxy $\mathbf{Q}_{\text{proxy}}=\bigcup_{i}\mathbf{Q}_i^{(T)}$, proxy anchor-based attention map for the current window is represented as follows:
\begin{align}
\mathbf{S}_{\text{proxy}} = \mathbf{Q}_\text{proxy}  \cdot \mathbf{K}_{\text{global}}^\top \in \mathbb{R}^{N \times (BN)}.
\end{align}
% \WJ{0801/10AM} % Q^proxy는 왜 위첨자임? 아래첨자로 첨자 위치 통일 해야함. K 는 K_global, j 로 하면서 그걸ㄹ ㅗ통일하던가해주셈.

The proxy is located at the center of the high-similarity embedding samples, aggregated from both inner- and outer-window regions. As a result, the proxy inherently balances the contributions of tokens across windows, allowing attention to be allocated more uniformly to semantically relevant regions, as shown in Fig.~\ref{fig:attention_ablation_viz}(b).
% This process ensures that each proxy is semantically grounded and robust, leading to more meaningful and stable attention interactions.
% 이걸로 attention stabilization 됐다는 근거는 Table어쩌구에 있다.

\subsection{Dynamic Normalization across Object Scales} % Robust attention normalization
\label{sec:dynamic_norm}

Our key-value extension allows each query to attend to tokens from global windows. While this enables broader contextual integration, it also increases the risk of attending to irrelevant or semantically mismatched tokens (i.e., negatives). This issue is strongly influenced by object scale: when a query corresponds to a small object or fine-grained region, the number of relevant positive tokens tends to be low. In such cases, attention is more easily dominated by irrelevant tokens, leading to inaccurate or noisy predictions.
% , as illustrated in Fig.~\ref{fig:attention_ablation_viz}.

To address this, we propose a dynamic normalization strategy that modulates and thresholds the similarity map in a scale-aware manner. The key idea is that when a query is associated with only a few positive tokens, as is often the case with small objects, the attention becomes more prone to being dominated by irrelevant or negative tokens, leading to degraded prediction quality. Our method adaptively tightens the normalization in such cases to suppress the effect of these irrelevant signals.
Conversely, when a query corresponds to a large object and is surrounded by many positive tokens, the normalization becomes more permissive to enhance their relative influence in the attention computation.
Additionally, the influence of irrelevant tokens increases with the number of outer-window tokens, amplifying noise in the attention computation. To handle this more effectively, our normalization strategy adjusts its scaling behavior based on the number of windows, suppressing the impact of such noise under larger global contexts.

Specifically, we introduce two adaptive variables, $\mathbf{u}$ and $\mathbf{w}$, as adaptive counterparts to the fixed hyperparameters in ProxyCLIP \cite{lan2024proxyclip} to modulate normalization accordingly. The attention score for the $i$-th query anchor is computed as:
% \WJ{0801/10AM} % 여기서 두개의 daptive variable소개한다그러면 앞에 hyperparameter가 늘어나느것처럼보이는데 앞에 감마베타를 대체할 dynamic term을 psi eta로 정의한다. 이런게 낫지않을까요?
% Specifically, we begin by defining a normalization equation as follows:
\begin{align}
\mathbf{Attn}_{i} &= \mathbf{w}_i \left( \mathbf{S}_{\text{proxy}} - \frac{\mathbf{u}}{NL} \sum_{j=1,...,NL} \left[ \mathbf{S}_{\text{proxy}} \right]_{ij} \right),
\label{eq:finalattn}
\end{align}
where $\mathbf{u}$ is a shifting variable dependent on the number of windows $L$, controlling the overall influence of extended tokens.
$\mathbf{u}$ is defined as:
% Here, $\mathbf{u}$ is a shifting variable dependent on the number of windows $L$, controlling the overall influence of extended tokens:
\begin{align}
\mathbf{u} &= 1 + \lambda_1 \log\bigl(1 + L),
\end{align}
where is a fixed coefficient. As $L$ increases, $\mathbf{u}$ grows, resulting in more conservative normalization to suppress noisy token influence.
Next, we define the scaling variable $\mathbf{w}_i$ based on the number of high confidence tokens $|\mathcal{P}_i|$ associated with each \textit{query}:
\begin{align}
\mathbf{w}_i = 1 + \frac{\lambda_2}{|\mathcal{P}_i|}.
\label{eq:dynamic_attention_scaling}
\end{align}
% \WJ{0801/10AM} % fixed coefficient 대신 fixed?small? constant? 이런게 나을듯.
A smaller $|\mathcal{P}_i|$, often observed in small objects, leads to stronger amplification of the most relevant tokens while suppress the irrelevant tokens, encouraging scale-aware attention allocation. Note that, $\lambda_1$ and $\lambda_2$ are shared across all datasets.
Finally, replacing $\mathbf{A}$ in Eq.~\eqref{eq:extended_visual_token} with $\mathbf{Attn}$ in Eq.~\eqref{eq:finalattn} yields the final visual tokens for prediction. The masking and softmax operations are then applied, as formulated in Eq.~\eqref{eq:proxy_attention}.

% \SE{08/01} % 위 수식에 대한 어떠한 rationale도 없는거죠? 왜 이 형태의 수식이 되어야한다.
%  그리고 어떠한 query 별로 attention이 adaptive 해야한다는 말은 존재하는데, 왜 이게 thresholding, scaling 형태가 되어야한다 것에 대한 설명은 부재합니다. <-- 필요해보임
% 그리고 중간에 글에서 "이렇게 하면 nomalization 이 노이즈 제거해준다" 이렇게 쓰지 말고 조금 더 자세히 써주세요. "일정 이하 similarity의 토큰에 할당되는 attention을 0으로 만들어서 영향을 줄인다" 뭐 이렇게라도 쓰던가.
% 그리고 softmax 후 attention probability 에 thresholding하고 scaling하는게 맞나요? 그러니까 attention의 합이 1이 될 필요가 없는 상황인지?

A key distinction of our method is a per-query normalization strategy, compared to the prior approach~\cite{lan2024proxyclip} that applies identical normalization across all windows and tokens. By adapting attention modulation at the level of each anchor token, our formulation enables fine-grained control over the influence of global context. This scale-aware behavior is particularly effective for balancing attention across varying object sizes. Furthermore, Dynamic Normalization copes with object scale variations, which differ among datasets, and thus removes the necessity of dataset-specific hyperparameters unlike recent methods~\cite{resclip, CASS, FreeCP}.

\section{Experiment}

\begin{table*}[t!]
\centering
\renewcommand{\arraystretch}{0.95}
\setlength{\tabcolsep}{4pt}
\small
\vspace{-0.2cm}
\begin{tabular}{l|c|c|c|cccc|cccc|c}
\hline
\multirow{2}{*}{Model} & \multirow{2}{*}{Train} & \multirow{2}{*}{\parbox[c]{1.5cm}{\centering External\\Model}} & \multirow{2}{*}{\parbox[c]{2cm}{\centering \textsuperscript{*}DS\\Hyperparameter}} & \multicolumn{4}{c|}{With Background} & \multicolumn{4}{c|}{Without Background} & \multirow{2}{*}{Avg} \\
& & & & V21 & PC60 & C-Obj & V20 & PC59 & C-Stf & City & ADE & \\
\hline
% CLIP-DINOiser
\multicolumn{1}{l}{\cellcolor[gray]{0.93}\textbf{CLIP-DINOiser Setting}} & 
\multicolumn{12}{l}{\cellcolor[gray]{0.93}\footnotesize 448 Image Resize, 448 Crop size, 224 Stride, Without rename trick\textsuperscript{\dag}} \\
GroupViT ~\cite{xu2022groupvit} & \checkmark & - & \ding{56} & 50.4 & 18.7 & 27.5 & 79.7 & 23.4 & 15.3 & 11.1 & 9.2 & 29.4 \\
ReCo ~\cite{reco} & \checkmark & - & \ding{56} & 25.1 & 19.9 & 15.7 & 57.7 & 22.3 & 14.8 & 21.6 & 11.2 & 23.5 \\
CLIP-DIY ~\cite{clip_diy} & \ding{56} & - & \ding{56} & 59.9 & 19.7 & 31.0 & 79.7 & 19.8 & 13.3 & 11.6 & 9.9 & 30.6 \\
TCL ~\cite{TCL} & \checkmark & - & \ding{56} & 55.0 & 30.4 & 31.6 & \textbf{83.2} & 33.9 & 22.4 & 24.0 & 17.1 & 37.2 \\
CLIP-DINOiser  ~\cite{clip-dinoiser} & \checkmark & DINO & \ding{56} & 62.2 & 32.4 & 35.0 & 80.2 & 35.9 & 24.6 & 31.7 & \textbf{20.0} & \underline{40.3} \\
\textbf{ProxyCLIP + GLA\textsuperscript{\ddag}} & \ding{56} & DINO & \ding{56} & \textbf{63.4} & \textbf{35.8} & \textbf{37.3} & \underline{80.4} & \textbf{39.4} & \textbf{26.3} & \textbf{33.5} & \underline{19.1} & \textbf{41.9} \\
\hline
% ClearCLIP
\multicolumn{1}{l}{\cellcolor[gray]{0.93}\textbf{ClearCLIP Setting}} & 
\multicolumn{12}{l}{\cellcolor[gray]{0.93}\footnotesize 448 Image Resize, 448 Crop size, 224 Stride, Without rename trick} \\
ClearCLIP ~\cite{clearclip}  & \ding{56} & - & \ding{56} & 51.8 & 32.6 & 33.0 & \textbf{80.9} & 35.9 & 23.9 & 30.0 & 16.7 & \underline{38.1} \\
\textbf{ClearCLIP + GLA} & \ding{56} & - & \ding{56} & \textbf{55.6} & \textbf{33.2} & \textbf{34.6} & \underline{79.7} & \textbf{36.5} & \textbf{24.6} & \textbf{32.7} & \textbf{17.4} & \textbf{39.3} \\
\hline
% ProxyCLIP
\multicolumn{1}{l}{\cellcolor[gray]{0.93}\textbf{ProxyCLIP Setting}} & 
\multicolumn{12}{l}{\cellcolor[gray]{0.93}\footnotesize 448 Image Resize, 336 Crop size, 112 Stride, With rename trick on Background Classes} \\
ProxyCLIP ~\cite{lan2024proxyclip} & \ding{56} & DINO & \ding{56} & 61.3 & 35.3 & 37.5 & 80.3 & 39.1 & 26.5 & 38.1 & 20.2 & \underline{42.3} \\
% FSA ~\cite{FSA}  & \ding{56} & DINO & \checkmark & 63.7 & 36.1 & 38.0 & 82.3 & 39.9 & 27.0 & 38.8 & 20.5 & 43.3 \\
\textbf{ProxyCLIP + GLA} & \ding{56} & DINO & \ding{56} & \textbf{63.2} & \textbf{35.8} & \textbf{37.7} & \textbf{81.5} & \textbf{39.7} & \textbf{26.8} & \textbf{38.5} & \textbf{20.3} & \textbf{42.9} \\
\hline
% SCLIP
\multicolumn{1}{l}{\cellcolor[gray]{0.93}\textbf{SCLIP Setting}} & 
\multicolumn{12}{l}{\cellcolor[gray]{0.93}\footnotesize 336 Image Resize, 224 Crop size, 112 Stride, With rename trick} \\
SCLIP ~\cite{sclip} & \ding{56} & - & \ding{56} & 59.1 & 30.4 & 30.5 & 80.4 & 34.2 & 22.4 & 32.2 & 16.1 & \underline{38.2} \\
\textbf{SCLIP + GLA} & \ding{56} & - & \ding{56} & \textbf{59.8} & \textbf{32.0} & \textbf{32.8} & \textbf{80.5} & \textbf{35.1} & \textbf{24.2} & \textbf{35.1} & \textbf{18.0} & \textbf{39.8} \\
\hline
LaVG  ~\cite{lavg} & \ding{56} & - & \ding{56} & 62.1 & 31.6 & 34.2 & 82.5 & 34.7 & 23.2 & 26.2 & 15.8 & 38.8 \\
NACLIP ~\cite{naclip} & \ding{56} & - & \ding{56} & 58.9 & 32.2 & 33.2 & 79.7 & 35.2 & 23.3 & 35.5 & 17.4 & 39.4 \\
ResCLIP  ~\cite{resclip} & \ding{56} & - & \checkmark & 61.1 & 33.5 & 35.0 & \underline{86.0} & 36.8 & 24.7 & 35.9 & 18.0 & 41.4 \\
FreeCP  ~\cite{FreeCP} & \ding{56} & - & \checkmark & 65.8 & 35.3 & 37.2 & 84.3 & 38.0 & 24.9 & 33.3 & 18.4 & 42.2 \\
DIH-CLIP  ~\cite{DIH-CLIP} & \ding{56} & - & \ding{56} & 64.2 & 36.0 & 37.4 & 84.9 & 39.7 & 24.5 & 40.2 & 19.6 & 43.3 \\
FLOSS ~\cite{FLOSS} & \ding{56} & - & \ding{56} & - & - & - & 80.2 & 35.9 & 23.6 & 37.0 & 18.4 & - \\
CASS  ~\cite{CASS} & \ding{56} & DINO & \checkmark & 65.8 & \textbf{36.7} & \textbf{37.8} & \textbf{87.8} & \textbf{40.2} & 26.7 & 39.4 & \underline{20.4} & \textbf{44.4} \\
% \textbf{ProxyCLIP + GLA} & \ding{56} & DINO & \ding{56} & \textbf{66.2} & \underline{36.2} & \textbf{37.8} & 83.7 & \underline{40.1} & \textbf{27.2} & \textbf{41.2} & 
% \underline{20.0} & \underline{44.0} \\
\textbf{ProxyCLIP + GLA (Best\textsuperscript{\S}}) & \ding{56} & DINO & \checkmark & \textbf{66.7} & \underline{36.3} & \underline{37.7} & 84.7 & \textbf{40.2} & \textbf{27.2} & \textbf{41.2} & \textbf{20.5} & 44.3 \\
\hline
\textbf{ProxyCLIP + GLA} & \ding{56} & DINO & \ding{56} & \underline{66.3} & 36.1 & \underline{37.7} & 84.2 & \underline{39.9} & \underline{26.9} & \underline{40.8} & 20.0 & \underline{44.0} \\
% \textbf{ProxyCLIP + GLA - Multi Layer Feat} & \ding{56} & DINO & \ding{56} & \textbf{67.0} & \underline{36.5} & \textbf{37.8} & 84.2 & \underline{40.5} & \textbf{27.4} & \textbf{42.3} & \underline{20.3} & \underline{44.6} \\
\hline
\end{tabular}
\vspace{-0.1cm}
\caption{
Open-vocabulary semantic segmentation results on 8 datasets using CLIP ViT-B/16. 
\textsuperscript{*}DS Hyperparameter: dataset-specific hyperparameters (not used in Ours). 
\textsuperscript{\dag}Rename trick: multiple textual prompts per class. 
\textsuperscript{\ddag}GLA: Adaptation for ClearCLIP~\cite{clearclip}, SCLIP~\cite{sclip}, and ProxyCLIP~\cite{lan2024proxyclip}.
\textsuperscript{\S} Best: optionally uses dataset-specific hyperparameters (\textbf{u}, \textbf{w}), by manual tuning.
}
\label{tab:sota_ovs_comparison}
\vspace{-0.2cm}
\end{table*}

% \textcolor{blue}{s}
% CASS (CVPR2025에서 베껴 옴)
\subsection{Experimental Settings} 
\label{sec:experimental_settings}
\textbf{Datasets and Evaluation Metrics}
We evaluate the proposed model on eight semantic segmentation benchmarks. 
We note that five datasets~(Pascal VOC21~\cite{pascalvoc}, Pascal Context60~\cite{pascalcontext}, COCO-Object~\cite{coco}, ADE20K~\cite{ade20k}, and Cityscapes~\cite{cordts2016cityscapes}) include an explicit background class, while other datasets~(Pascal VOC20~\cite{pascalvoc}, Pascal Context59~\cite{pascalcontext}, and COCO-Stuff 164K~\cite{coco}) only contain object classes.
% Five datasets include an explicit background class: Pascal VOC21~\cite{pascalvoc}, Pascal Context60~\cite{pascalcontext}, COCOObject~\cite{coco}, ADE20K~\cite{ade20k}, and Cityscapes~\cite{cordts2016cityscapes}, while three are background free: Pascal VOC20~\cite{pascalvoc}, Pascal Context59~\cite{pascalcontext} , and COCOStuff 164K~\cite{coco}.
In the same order, they comprise 21, 60, 81, 150, 19, 20, 59, and 171 semantic classes, respectively. 
Following standard practice, we report performance using mean Intersection‑over‑Union~(mIoU).

\noindent \textbf{Implementation Details.} % 원본: The detailed settings are same with existing training-free open-vocabulary semantic segmentation models~\cite{naclip, lavg, lan2024proxyclip, cliptrase, sclip}. For the text prompts, however, we adopted the SCLIP~\cite{sclip} setting for fair comparison, specifically, it provides multiple prompts for some classes. We use frozen CLIP (ViT‑B/16) and DINO (ViT‑B/8) backbones as our VFM. For the image preprocessing, we resize images to a 336 pixels for the shorter side~(560 pixels for Cityscapes). Following the trained resolution of CLIP, we apply 224×224 sliding‑window crops with a 112 pixels stride, and run all experiments on a single NVIDIA Titan RTX GPU. We set the proxy-anchor-related hyperparameters as follows: $\rho = 0.6$ and the number of proxy generation steps to 2. For Dynamic Normalization, we used $\lambda_1 = 0.3$ and $\lambda_2 = 30$.
Our settings follow existing training-free open-vocabulary semantic segmentation models~\cite{clip-dinoiser, clearclip, lan2024proxyclip, sclip}, including text prompt policy, background threshold, image resizing, sliding window crop size, and stride (details in the Appendix~\ref{sec:setting_detail}). We use frozen CLIP (ViT-B/16) to extract features and DINO (ViT-B/8) as VFM, similar to ProxyCLIP~\cite{lan2024proxyclip}. 
Meanwhile, the query and key tokens from CLIP can also be utilized to form attention maps, replacing VFM features. ClearCLIP~\cite{clearclip} and SCLIP~\cite{sclip} adopt this strategy, and our framework can similarly support it. However, in this case, a separate query smoothing step and hyperaprater tuning is necessary, as detailed in the Appendix~\ref{sec:query_smoothing}. All experiments are conducted on a single NVIDIA Titan RTX GPU. Proxy-anchor hyperparameters are set to $\rho = 0.6$ with 2 proxy generation steps, and Dynamic Normalization uses $\lambda_1 = 0.3$ and $\lambda_2 = 30$.

\subsection{Comparison with State-of-the-Art Methods}
\textbf{Quantitative Results.}
% 원본: 
% As shown in Tab.~\ref{tab:sota_ovs_comparison}, we compare our method against existing OVSS approaches.
% % To highlight robustness of the proposed method across various datasets, we annotate the ‘$\mathfrak{R}$’ column with a $\checkmark$ for methods that use a unified set of hyperparameters across datasets and an \xmark \ for those that require dataset-specific tuning.
% % Methods marked with an \xmark \ indicate a lack of generalizability, as they rely on individually tuned hyperparameters for each dataset.
% Our method demonstrates consistent performance without any dataset-specific hyperparameter tuning, showcasing strong robustness. 
% Compared to ProxyCLIP~\cite{lan2024proxyclip}, ours achieves an average mIoU improvement 3.6\%. 
% % Moreover, despite CASS~\cite{CASS} leveraging dataset-specific hyperparameter tuning, our method achieves comparable performance without such tailoring. 
% These results suggest that our approach can be effectively and easily deployed across diverse datasets and domains without manual intervention.
As shown in Tab.~\ref{tab:sota_ovs_comparison}, we compare our method with existing open-vocabulary semantic segmentation (OVSS) approaches and report its performance under each setting for a fair evaluation.
Our method achieves an average mIoU gain of 1.6\% under the CLIP-DINOiser setting, demonstrating clear advantages over training-based models. When integrated with ClearCLIP~\cite{clearclip}, ProxyCLIP~\cite{lan2024proxyclip}, and SCLIP~\cite{sclip}, it further yields improvements of 1.2\%, 0.6\%, and 1.6\%, respectively, highlighting its adaptability on any baseline model.
In contrast to recent training-free OVSS methods~\cite{FSA, resclip, FreeCP, CASS}, which depend on dataset-specific hyperparameters to enhance performance at the cost of generality, our approach attains an average mIoU of 44.0\%, outperforming existing models without tuning. When dataset-specific hyperparameters are optionally applied, our model achieves even higher scores, comparable to all baseline methods.
These results indicate that our approach is both effective and generalizable, making it readily applicable across diverse datasets.

% Related-Work
\begin{figure*}
    \centering
    \vspace{-0.2cm}
    \includegraphics[width=0.95\textwidth]{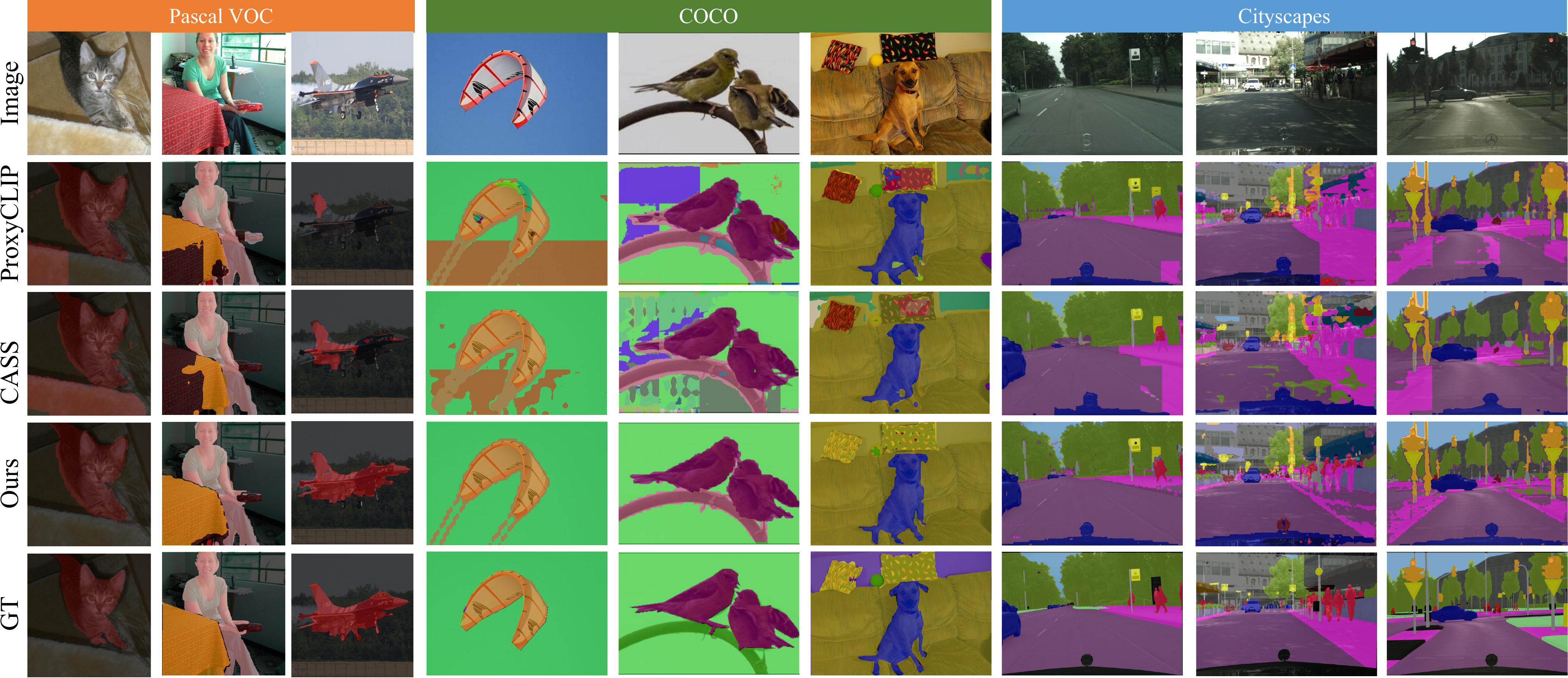}
    \vspace{-0.2cm}
    \caption{
        Qualitative results among ProxyCLIP~\cite{lan2024proxyclip}, CASS~\cite{CASS}, Ours on Pascal VOC21~\cite{pascalvoc}, COCOstuff~\cite{coco}, and Cityscapes~\cite{cordts2016cityscapes}. 
    }
    \label{fig:Visualization}
    \vspace{-0.35cm}
\end{figure*}

%원본: Fig.~\ref{fig:Visualization} presents qualitative results between ours and recent state-of-the-art approaches on the Pascal VOC21, COCO-Stuff164k, and Cityscapes datasets. Compared to the baselines, our method consistently produces segmentation maps with fewer misclassified pixels. Notably, our key-value extension combined with proxy queries effectively mitigates the grid-like artifacts frequently observed in ProxyCLIP~\cite{lan2024proxyclip}. These results demonstrate that ours can leverage contextual cues from global windows, leading to consistent predictions across windows. For instance, in the rightmost Cityscapes example, ProxyCLIP incorrectly predicts the center of the road as a sidewalk due to residual grid patterns. In comparison, our model clearly captures the global context and delineates the boundary between road and sidewalk without introducing noisy or inconsistent predictions.
\noindent \textbf{Qualitative Results.}
Fig.~\ref{fig:Visualization} shows qualitative comparisons with recent state-of-the-art methods on Pascal VOC21~\cite{pascalvoc}, COCO-Stuff164k~\cite{coco}, and Cityscapes~\cite{cordts2016cityscapes}. Our method generates cleaner segmentation maps with fewer misclassified pixels and effectively suppresses the grid-like artifacts commonly seen in ProxyCLIP~\cite{lan2024proxyclip} and CASS~\cite{CASS}. By leveraging contextual cues from global windows, our model produces more consistent predictions and preserves clear object boundaries. For example, in the rightmost scene in Cityscapes, ProxyCLIP misclassifies the road center as a sidewalk, whereas ours accurately distinguishes the road and sidewalk regions without noisy or inconsistent outputs.

\section{Further Analysis}

%%%%%%%%%%%%%%%%%%%%%%%%%%%%%%%%%%%%%%%%%%%%%%%%%%%%%%%%%%%%
%%%%%%%%%%%%%%%%%%%%%%%%%%%%%%%%%%%%%%%%%%%%%%%%%%%%%%%%%%%%

\begingroup
\renewcommand{\arraystretch}{1.0}  % row 높이
\begin{table}[t]
\centering
\footnotesize  % 글자 크기

% \begin{tabularx}{\columnwidth}{
% c|
% >{\centering\arraybackslash}X|
% >{\centering\arraybackslash}X|
% >{\centering\arraybackslash}p{0.12\columnwidth}|
% >{\centering\arraybackslash}p{0.12\columnwidth}|
% c}
% \toprule
% & \multirow{2}{*}{\textbf{KVE}} & \multirow{2}{*}{\textbf{Proxy}} & \multicolumn{2}{c|}{\textbf{Norm}} & \multirow{2}{*}{\textbf{mIoU}} \\
% \cline{4-5}
%  & & & \textbf{Best} & \textbf{Dyna} & \\
% \midrule
% (a) &  &  &  & & 30.8    \\
% % (a) &  &  & & & 42.5    \\ % proxyclip
% \midrule
% % (b) & \checkmark &  &  &  & 25.6 \\
% % (d) & \checkmark & \checkmark &  & & 26.1 \\
% % (d) & \checkmark &  & \checkmark & & 38.1 \\
% % (d) & \checkmark & \checkmark & \checkmark & & 44.2 \\
% % (e) &  & \checkmark & & \checkmark & 43.0 \\
% (b) & \checkmark &  & & \checkmark &  43.1 \\
% (c) & & \checkmark  & & \checkmark &  43.0 \\
% \midrule
% (d) & \checkmark & \checkmark & \checkmark & & 44.3 \\
% (e) & \checkmark & \checkmark  & & \checkmark & 44.0 \\
% \bottomrule
% \end{tabularx}

\begin{tabularx}{\columnwidth}{c|c|c|>{\centering\arraybackslash}X|>{\centering\arraybackslash}X|>{\centering\arraybackslash}X|c}
\toprule
& \multirow[c]{2}{*}{\textbf{KVE}} & \multirow[c]{2}{*}{\textbf{Proxy}} & \multicolumn{3}{c|}{\textbf{Norm}} & \multirow[c]{2}{*}{\textbf{mIoU}} \\
\cline{4-6}
 & & & \textbf{Fixed} & \textbf{Best} & \textbf{Dyna} & \\
\midrule
(a) &  &  &  &  & & 30.8    \\
\midrule
% (b) &  &  & \checkmark &  &  & 42.5 \\
(b) & \checkmark &  &  &  & \checkmark &  43.1 \\
(c) & & \checkmark  &  &  & \checkmark &  43.0 \\
\midrule
(d) & \checkmark & \checkmark & \checkmark & \checkmark &  & 44.3 \\
(e) & \checkmark & \checkmark  &  & \checkmark & \checkmark & 44.0 \\
\bottomrule
\end{tabularx}

% \begingroup
% \renewcommand{\arraystretch}{1.0}  % row 높이
% \begin{table}[t]
% \centering
% \footnotesize  % 글자 크기
% \begin{tabularx}{\columnwidth}{c|c|c|>{\centering\arraybackslash}X|>{\centering\arraybackslash}X|>{\centering\arraybackslash}X|c}
% \hline
% & \multirow[c]{2}{*}{\textbf{KVE}} & \multirow[c]{2}{*}{\textbf{Proxy}} & \multicolumn{3}{c|}{\textbf{Norm}} & \multirow[c]{2}{*}{\textbf{mIoU}} \\
% \cline{4-6}
%  & & & \textbf{Fixed} & \textbf{Best} & \textbf{Dyna} & \\
% \hline
% (a) & & & & & & 30.8 \\
% \hline
% (b) &               &   & \checkmark &              &   & 42.5 \\
% \hline
% (c) & \checkmark &   & \checkmark &  &  & 38.1 \\
% % (d) & \checkmark &   &  & \checkmark &  & 43.8 \\
% (d) & \checkmark &   &  &  &  \checkmark & 43.2 \\
% \hline
% (e) &  &  \checkmark &  \checkmark &  &  &  42.9 \\
% % (g) &  & \checkmark  &  & \checkmark &  & 44.2 \\
% (f) &  &  \checkmark &  &  &  \checkmark  &  43.0 \\
% \hline
% (g) & \checkmark & \checkmark & \checkmark &  &  & 39.5 \\
% (h) & \checkmark & \checkmark  &  & \checkmark &  & 44.3 \\
% (i) & \checkmark & \checkmark  &  &  & \checkmark & 44.0 \\
% \hline
% \end{tabularx}

\vspace{-0.3cm}
\caption{Ablation study of each component in our method. KVE: Key-Value Extension, Proxy: Proxy Anchor-based Attention, Norm: Normalization (Best, Dynamic). Dataset-specific tuning is applied for `Best' normalization configuration, with detailed results in Tab.~\ref{tab:sota_ovs_comparison}.}
\label{tab:ablation_study}
\vspace{-0.7cm}
\end{table}
\endgroup

\vspace{-0.1cm}
\subsection{Ablation Study}
\vspace{-0.1cm}

To validate the effectiveness of each component, we conduct an ablation study based on five experimental configurations, as summarized in Tab.~\ref{tab:ablation_study}. We begin with a baseline~(a) that computes attention maps using features from DINO, restricted to inner-window tokens only, without any form of normalization. Comparing (a) and (b), we observe a clear performance improvement when applying key-value extension along with dynamic normalization. This demonstrates the importance of incorporating global context and stabilizing the attention computation. Next, comparing (b) and (e), the introduction of proxy-based attention in (e) leads to an additional performance gain. This improvement stems from the proxy’s ability to produce more semantically grounded and stable attention distributions, especially in outer-window regions. We also compare (c) and (e) to isolate the contribution of the key-value extension. The results show that incorporating key-value tokens from outer windows yields a performance improvement, confirming that these extended tokens provide valuable contextual cues for disambiguating local predictions. Finally, experiment (d) reports results using dataset-specific hyperparameters, which were searched to maximize performance for each dataset. When comparing (d) and our method (e), we find that the performance gap is minimal, indicating that our adaptive normalization approach effectively generalizes across datasets without the need for manual tuning.

\subsection{Effect of Attention Stabilization}

We visualize the effect of each method on attention maps in Fig.~\ref{fig:attention_ablation_viz}. In Fig.~\ref{fig:attention_ablation_viz}(a), the attention maps after applying Key-Value Extension show that attention initially concentrates on a few tokens within the inner-window region, indicating that the model struggles to distribute attention across object boundaries and capture object-level semantics beyond individual windows. Over time, the strong attention weights originally confined to inner-window positives gradually spread to outer regions, leading to a holistic, object-level attention allocation. 
In Fig.~\ref{fig:attention_ablation_viz}(b), after applying Proxy-Attention, attention extends across the entire object regardless of window boundaries, showing that proxy queries effectively stabilize attention. 
Subsequently, in Fig.~\ref{fig:attention_ablation_viz}(c), introducing Dynamic Normalization suppresses irrelevant tokens and focuses attention on meaningful object regions, highlighting its role in refining attention distribution.

% 원본
% \subsection{Analysis of Dynamic Normalization}
% \textcolor{blue}{
% % As shown in Fig.~\ref{fig:attention_ablation_viz}(c), we qualitatively analyze the effect of Dynamic Normalization. Proxy-Attention reduces window biases, suppresses irrelevant tokens, and concentrates attention on meaningful object regions, demonstrating its ability to refine attention distribution. 
% For quantitative analysis, we evaluate Dynamic Normalization on Cityscapes \cite{cordts2016cityscapes}. Fixed attention scaling exhibits a trade-off. Low attention scaling (\textbf{w} = 1, 2, 3, 4) favors large objects at the expense of small objects, achieving peak performance at 40.6 mIoU at (\textbf{w} = 5), while high attention scaling (\textbf{w} = 6, 7, 8) harms large-object segmentation due to diminished global context. Dynamic Normalization adaptively handles varying object scales, combining the benefits of both extremes and achieving superior performance of 40.8 mIoU, confirming its effectiveness. }

\subsection{Object Scale via high confidence Tokens}
\label{sec:object_scale}

\begin{figure}[t]
    \centering
    \vspace{-0.3cm}
    \includegraphics[width=\columnwidth]{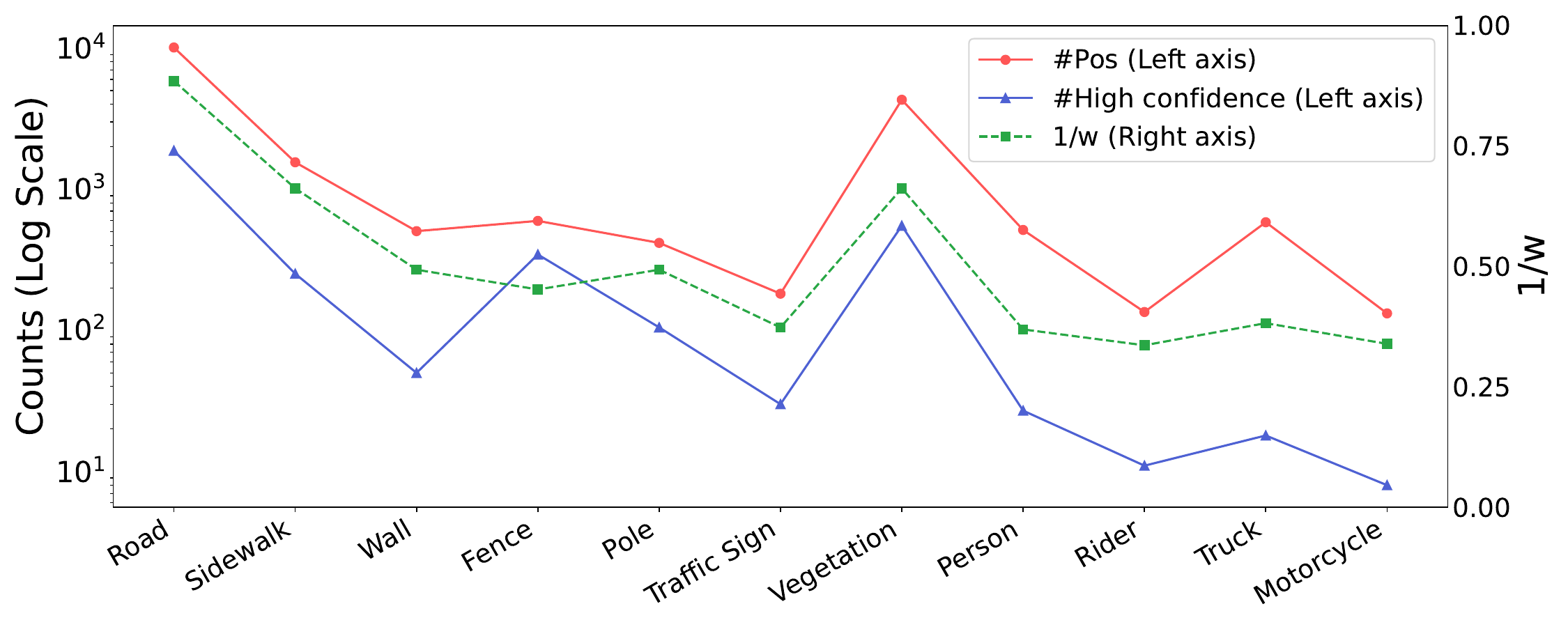}
    \vspace{-0.7cm}
    \caption{
        Class-wise object scale and 1/\textbf{w} in Cityscapes
    }
    \label{fig:cityscapes_object_scale}
    \vspace{-0.2cm}
\end{figure}

We conduct experiments to validate the assumption that high confidence tokens are correlated with object scale in Dynamic Normalization. During proxy construction, we use the number of high confidence tokens, $|\mathcal{P}_i|$, as an estimate of object scale and incorporate it into the attention scaling factor, as defined in Eq.~\ref{eq:dynamic_attention_scaling}.
In Fig.~\ref{fig:cityscapes_object_scale}, we analyze whether the number of high confidence tokens aligns with the ground-truth object scale on the Cityscapes~\cite{cordts2016cityscapes}, which contains objects of diverse sizes. After key-value extension, we measure both the number of patch tokens (\#Pos) and high confidence tokens (\#High confidence) for each class per image, and report the class-wise average attention scaling factor $\mathbf{w}$, which is inversely proportional to \#High confidence.
We observe consistent trends across all three metrics: large objects (e.g., Road and Terrain) exhibit a greater number of high confidence tokens and correspondingly lower attention scaling factors, whereas small objects (e.g., Person and Rider) contain fewer tokens and thus receive higher attention scaling.

\begin{figure}[t]
    \centering
    \vspace{-0.2cm}
    \includegraphics[width=\columnwidth]{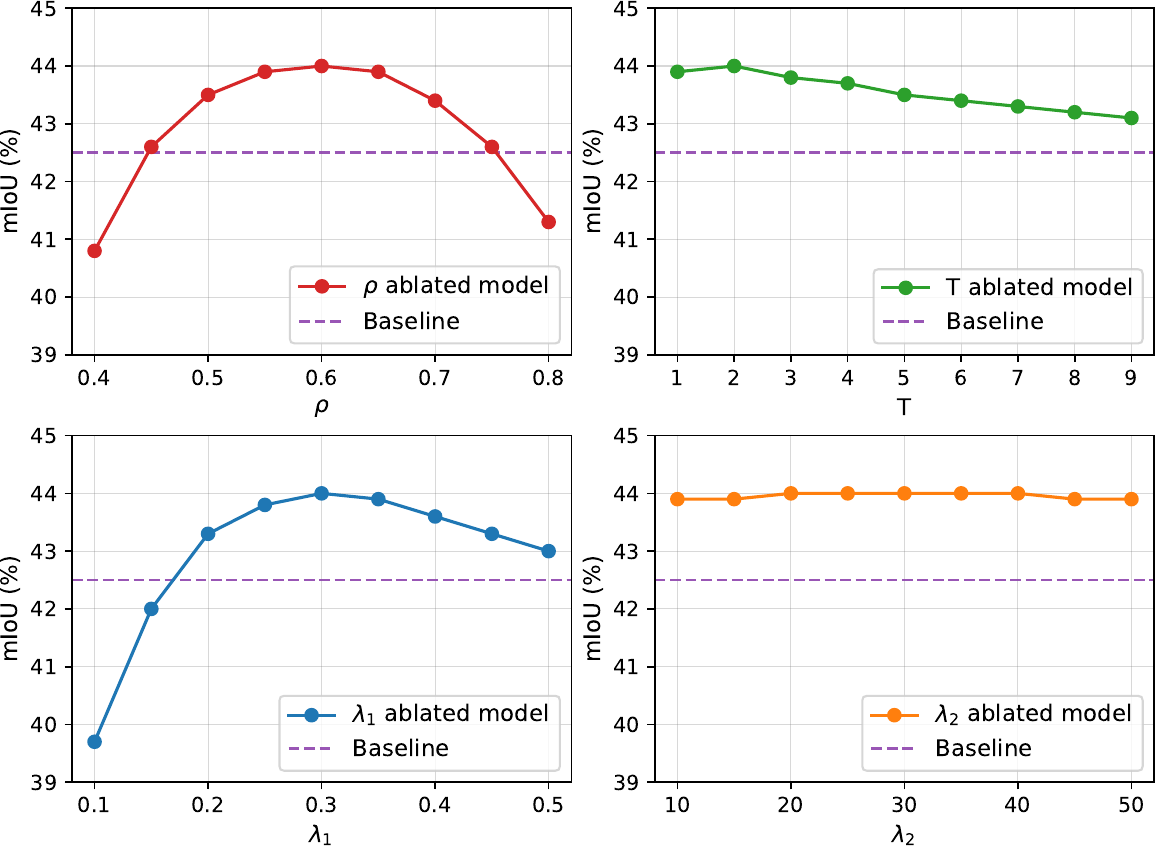}
    \vspace{-0.6cm}
    \caption{
        Hyperparameter sensitivity experiment.
    }
    \label{fig:hyperparameter_effects}
    \vspace{-0.6cm}
\end{figure}

\subsection{Hyperparameter Sensitivity}
We leverage four hyperparameters, $\rho$, T, $\lambda_1$, and $\lambda_2$, and conduct a sensitivity analysis to examine their effect (Fig. ~\ref{fig:hyperparameter_effects}). Firstly, $\rho$ controls the selection of high confidence tokens in the cosine similarity scale. If $\rho$ is too low, negatives may be fused with positives, causing semantic collapse in proxy anchors and degrading performance, while too high $\rho$ prevents proxies from aggregating sufficient positives, also reducing performance. The repetition step T in proxy anchor construction slightly affects performance, as a higher T can lead to semantic collapse.
In Dynamic Normalization, $\lambda_1$ determines the attention masking threshold. A low $\lambda_1$ lowers the threshold, causing fewer tokens to be masked and more dispersed attention, whereas a higher $\lambda_1$ increases masking and helps to focus attention effectively. In contrast, $\lambda_2$ has minimal impact on performance, revealing its robustness. Overall, all hyperparameters exhibit a broad range to consistently outperform the baseline model.

\begingroup
\renewcommand{\arraystretch}{0.9}  % row 높이

% -------------------------------
% 1. CLIP Table
% -------------------------------
% \begin{table}[t]
% \centering
% \scriptsize  % 글자 크기
% \begin{tabularx}{\columnwidth}{
% >{\centering\arraybackslash}p{1.4cm}|
% >{\centering\arraybackslash}p{1.4cm}|  % Model 열 고정폭 중앙정렬
% >{\centering\arraybackslash}X|
% >{\centering\arraybackslash}X|
% >{\centering\arraybackslash}X|
% >{\centering\arraybackslash}X
% }
% \hline
% \textbf{CLIP} & \textbf{Model} & \textbf{VOC21} & \textbf{PC59} & \textbf{C-Stf} & \textbf{Avg.} \\
% \hline
% \multirow{3}{*}{ViT-B/32} 
%  & ProxyCLIP & 57.3 & 35.2 & 23.6 & 38.7 \\
%  & CASS      & 58.2 & 36.5 & 24.4 & 39.7 \\
%  \cline{2-6}
%  % \cmidrule(lr){2-6}
%  & Ours      & \textbf{61.8} & \textbf{37.1} & \textbf{25.2} & \textbf{41.4} \\
% \hline
% \multirow{3}{*}{ViT-L/14} 
%  & ProxyCLIP & 60.7 & 38.2 & 26.2 & 41.7 \\
%  & CASS      & 62.1 & \textbf{39.1} & 26.3 & 42.5 \\
%  \cline{2-6}
%  % \cmidrule(lr){2-6}
%  & Ours      & \textbf{63.5} & \textbf{39.1} & \textbf{26.7} & \textbf{43.1} \\
% \hline
% \end{tabularx}
% \vspace{-0.2cm}
% \caption{Segmentation results of different CLIP backbones.}
% \label{tab:clip_backbone_ablation}
% \end{table}
\begin{table}[t]
\vspace{-0.2cm}
\centering
\scriptsize
\renewcommand{\arraystretch}{1.05}

\begin{tabularx}{\columnwidth}{
>{\centering\arraybackslash}p{1.4cm}|
>{\centering\arraybackslash}p{1.4cm}|
>{\centering\arraybackslash}X|
>{\centering\arraybackslash}X|
>{\centering\arraybackslash}X|
>{\centering\arraybackslash}X
}

\toprule
\textbf{CLIP} & \textbf{Model} & \textbf{V21} & \textbf{PC59} & \textbf{C-Stf} & \textbf{Avg.} \\
\midrule

\multirow{3}{*}{ViT-B/32}
 & ProxyCLIP & 57.3 & 35.2 & 23.6 & 38.7 \\
 & CASS      & 58.2 & 36.5 & 24.4 & 39.7 \\
\cmidrule(lr){2-6}
 & \cellcolor{gray!15}{Ours}
 & \cellcolor{gray!15}\textbf{61.8}
 & \cellcolor{gray!15}\textbf{37.1}
 & \cellcolor{gray!15}\textbf{25.2}
 & \cellcolor{gray!15}\textbf{41.4} \\

\midrule

\multirow{3}{*}{ViT-L/14}
 & ProxyCLIP & 60.7 & 38.2 & 26.2 & 41.7 \\
 & CASS      & 62.1 & \textbf{39.1} & 26.3 & 42.5 \\
\cmidrule(lr){2-6}
 & \cellcolor{gray!15}{Ours}
 & \cellcolor{gray!15}\textbf{63.5}
 & \cellcolor{gray!15}\textbf{39.1}
 & \cellcolor{gray!15}\textbf{26.7}
 & \cellcolor{gray!15}\textbf{43.1} \\

\bottomrule
\end{tabularx}

\vspace{-0.3cm}
\caption{Segmentation results of different CLIP backbones.}
\label{tab:clip_backbone_ablation}
\vspace{-0.2cm}
\end{table}

% -------------------------------
% 2. VFM Table
% -------------------------------
\begin{table}[t]
\vspace{-0.2cm}
\centering
\scriptsize
\begin{tabularx}{\columnwidth}{
>{\centering\arraybackslash}p{1.4cm}|
>{\centering\arraybackslash}p{1.4cm}|
>{\centering\arraybackslash}X|
>{\centering\arraybackslash}X|
>{\centering\arraybackslash}X|
>{\centering\arraybackslash}X
}

\toprule
\textbf{VFM} & \textbf{Model} & \textbf{V21} & \textbf{PC59} & \textbf{C-Stf} & \textbf{Avg.} \\
\midrule

\multirow{2}{*}{DINOv2 \cite{dinov2}} 
 & ProxyCLIP & 59.6 & 37.3 & 25.3 & 40.7 \\
 & \cellcolor{gray!15}+GLA 
   & \cellcolor{gray!15}\textbf{60.8} 
   & \cellcolor{gray!15}\textbf{37.8} 
   & \cellcolor{gray!15}\textbf{25.7} 
   & \cellcolor{gray!15}\textbf{41.4} \\
   
\midrule

\multirow{2}{*}{DINOv3 \cite{dinov3}} 
 & ProxyCLIP & 62.8 & 38.8 & 26.1 & 42.6 \\
 & \cellcolor{gray!15}+GLA 
   & \cellcolor{gray!15}\textbf{63.6} 
   & \cellcolor{gray!15}\textbf{38.9} 
   & \cellcolor{gray!15}\textbf{26.3} 
   & \cellcolor{gray!15}\textbf{42.9} \\
   
\bottomrule
\end{tabularx}
\vspace{-0.3cm}
\caption{Segmentation results of different VFM backbones.}
\label{tab:vfm_backbone_ablation}
\vspace{-0.5cm}
\end{table}
\endgroup

\subsection{Generalization Across Backbones}
\label{sec:generalization_across_backbones}
% 다른 TF-OVS에서도 CLIP의 종류를 바꿔서 적용하는 Ablation 실험을 진행했었다.
% 우리도 실험을 통해 CLIP의 종류에 상관없이 성능을 높일 수 있는지 점검한다. 
% Furthermore, in Table~\ref{tab:clip_backbone_ablation}, our approach demonstrates strong robustness to changes in the CLIP backbone. While other methods exhibit significant performance degradation when switching to a different CLIP variant, our model maintains stable performance with only marginal drops. This indicates that our method does not heavily rely on the specific characteristics of a particular CLIP model, but rather leverages a more generalizable and architecture-agnostic design. The relatively consistent performance across different CLIP variants highlights the inherent robustness and adaptability of our algorithm.
To assess the generalizability of our framework across varying CLIP backbone capacities, we evaluate performance using ViT-B/32 and ViT-L/14. As shown in Tab.~\ref{tab:clip_backbone_ablation}, ProxyCLIP and CASS exhibits a substantial performance drop of 4.4\%, 4.5\% in average mIoU when switching to the lower capacity ViT-B/32. In contrast, our method incurs only a 3.0\% decrease, demonstrating improved robustness to representational capacity degradation. This indicates that our model more effectively leverages the available features across diverse CLIP architectures, and is less susceptible to variations in visual encoder strength. Furthermore, in Fig.~\ref{tab:vfm_backbone_ablation}, we conduct backbone exchanges in VFM to demonstrate the adaptability of our module. Specifically, we replace the backbone from DINO ViT-B/16 to DINOv2-reg ViT-L/14 ~\cite{dinov2, register} and DINOv3 ViT-B/16 ~\cite{dinov3} to verify the generality of our method. In each case, our approach yields performance improvements of +0.7\% and +0.3\%, respectively. Additionally, our method is compatible with complementary techniques such as multi-layer feature fusion and text-prompt modifications, as discussed in the Appendix~\ref{sec:multi-layer_feature_fusion},~\ref{sec:cass_adaptation}. Owing to its generality, our framework can be readily integrated with new backbones and future improvements.

% \begingroup
% \setlength{\tabcolsep}{3pt} % Default value: 6pt
% \renewcommand{\arraystretch}{1.0} % Default value: 1
% \begin{table}[t]
% \centering
% \begin{tabular}{l|l|cccc}
% \toprule
% \textbf{CLIP type} & \textbf{Model} & \textbf{V21} & \textbf{PC60} & \textbf{City} & \textbf{ADE} \\
% \midrule
% CLIP ViT-L/14      & Baseline &           &           &             &           \\
% CLIP      & + Ours   &           &           &             &           \\
% \midrule
% CLIP ViT-H/ & Baseline &           &           &             &           \\
% OpenCLIP  & + Ours   &           &           &             &           \\
% \midrule
% CLIP ViT-B/32 & Baseline &           &           &             &           \\
% OpenCLIP  & + Ours   &           &           &             &           \\
% \midrule
% DFNCLIP   & Baseline &           &           &             &           \\
% DFNCLIP   & + Ours   &           &           &             &           \\
% \bottomrule
% \end{tabular}
% \caption{Comparison of mIoU and inference time under different stride settings.}
% \label{tab:clip_comparison}
% \end{table}
% \endgroup

\section{Conclusion \& Limitation}
In this paper, we propose GLA-CLIP, a training-free method for OVSS that resolves inconsistent predictions across sliding windows. Using Key-Value Extension, each query attends to globally aggregated tokens, mitigating semantic discrepancies. We further address window bias and small object neglect by Proxy Anchor and Dynamic Normalization. While the Key–Value Extension utilizes whole tokens from other windows, computational cost may increase. More research is needed to efficiently leverage high-quality tokens across different windows. While our method can be adopted on any backbone features, certain hyperparameters may need to be adjusted to achieve optimal performance. We hope future research will address both memory efficiency and hyperparameter generalization.

\section*{Acknowledgements}
This work was supported in part by MSIT/IITP (No. RS-2022-II220680, RS-2020-II201821, RS-2019-II190421, RS-2024-00459618, RS-2024-00360227, RS-2024-00437633, RS-2024-00437102, RS-2025-25442569), MSIT/NRF (No. RS-2024-00357729), and KNPA/KIPoT (No. RS-2025-25393280).

{
    \small
    \bibliographystyle{ieeenat_fullname}
    \bibliography{main} % main.bib에서 Main 논문만 인용
}

\appendix
\clearpage
\setcounter{page}{1}
\maketitlesupplementary

\section{Boundary Error Rate (BER)}
\label{sec:BER}

\subsection{Definition of BER}

In Fig.~\ref{fig: motivation BER}, to measure prediction inconsistencies caused by the sliding-window mechanism, we introduce the Boundary Error Rate (BER). BER measures how often adjacent pixel pairs across window boundaries, represented as \(\{p, q\}\) in the equation, share the same ground-truth label but receive different predictions.

\begin{equation}
\text{BER} = 
\frac{
\sum\limits_{(p, q) \in \mathcal{B}} \mathbf{1}[(y_p = y_q) \wedge (\hat{y}_p \ne \hat{y}_q)]
}{
\sum\limits_{(p, q) \in \mathcal{B}} \mathbf{1}[y_p = y_q]
} \times 100
\end{equation}

\noindent
\textbf{Notation:}
\begin{itemize}
    \item \( p, q \): A pair of adjacent pixels located across a sliding window boundary.
    \item \( \mathcal{B} = \{(p, q)\} \): The set of all adjacent pixel pairs across window boundaries.
    \item \( y_p, y_q \): Ground-truth labels of pixels \( p \) and \( q \).
    \item \( \hat{y}_p, \hat{y}_q \): Predicted labels of pixels \( p \) and \( q \).
    \item \( \mathbf{1}[\cdot] \): Indicator function.
\end{itemize}

The denominator counts the number of adjacent pixel pairs across window boundaries that share the same ground-truth label. The numerator counts how many of those are predicted as different classes. BER reflects the proportion of inconsistencies.

\subsection{Complementarity of BER}

\begin{figure}[H]
    \centering
    \vspace{-0.2cm}
    \includegraphics[width=\columnwidth]{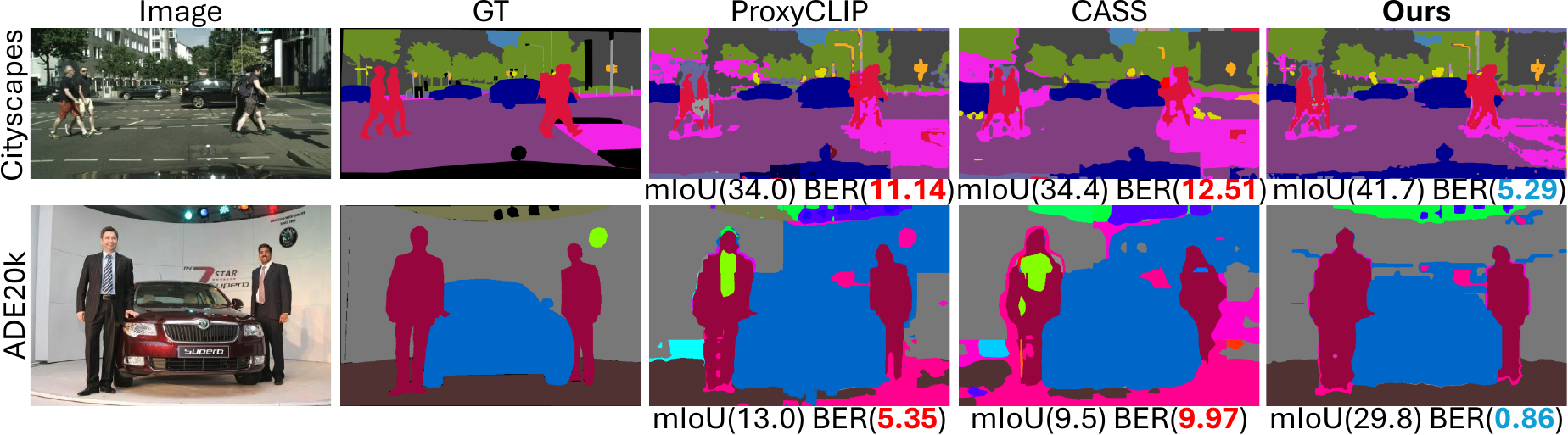}
    \vspace{-0.6cm}
    \caption{
        Qualitative Results with Sample-wise mIoU and BER.
    }
    \label{fig:BER vis}
    \vspace{-0.2cm}
\end{figure}

BER captures sliding-window grid artifacts overlooked by mIoU. In Fig.~\ref{fig:BER vis}, ours shows better sample mIoU than ProxyCLIP~\cite{lan2024proxyclip} \& CASS~\cite{CASS} and also produces natural masks with fewer grid artifacts, which BER correctly reflects. At dataset-level, gains in both mIoU \& BER shows accurate pixel classification and visually coherent segmentation.

\vspace{-0.1cm}

\section{Desciption of each setting}
\label{sec:setting_detail}

\subsection{CLIP-DINOiser Setting}
We follow CLIP-DINOiser ~\cite{clip-dinoiser} setting in Tab.~\ref{tab:sota_ovs_comparison} to conduct fair comparison. We resize input images with a short side of 448 and perform sliding-window inference with a 448 $\times$ 448 window and 224 stride for all dataset. 
We do not perform rename trick on each classnames and use only the standard ImageNet prompts following ~\cite{xu2022groupvit, maskclip}. Additionally, naive CLIP-DINOiser setting does not offer the background threshold unlike ClearCLIP ~\cite{clearclip, sclip, lan2024proxyclip}, we follow the background threshold of ClearCLIP to identify background in Pascal VOC21 ~\cite{pascalvoc} , Pascal Context60 ~\cite{pascalcontext} , COCO-Object ~\cite{coco}. 

\subsection{ClearCLIP Setting}
We follow ClearCLIP ~\cite{clearclip} setting in Tab.~\ref{tab:sota_ovs_comparison} to conduct fair comparison. We resize input images with a short side of 448 and perform sliding-window inference with a 448 $\times$ 448 window and 224 stride for all dataset. 
We do not perform rename trick on each classnames and use only the standard ImageNet prompts following ~\cite{xu2022groupvit, maskclip}. The background threshold for Pascal VOC21 ~\cite{pascalvoc}, Pascal Context60 ~\cite{pascalcontext}, COCO-Object ~\cite{coco} is assigned to 0.5, 0.15, 0.4, for each.

\subsection{ProxyCLIP Setting}
In ProxyCLIP ~\cite{lan2024proxyclip} setting in Tab ~\ref{tab:sota_ovs_comparison}, 
We resize the images to accommodate varying dataset specification: a shorter side of 336 pixels for PASCAL ~\cite{pascalvoc, pascalcontext} and COCO ~\cite{coco} datasets and 448 pixels for Cityscapes ~\cite{cordts2016cityscapes} and ADE20K ~\cite{ade20k} datasets. We adopt a sliding-window strategy with a 336 $\times$ 336 window and 112 $\times$ 112 stride. For the background class, rather than directly using the text prompt “background”, we employ a renaming strategy in which multiple class names associated with background semantics are grouped and used as substitutes. This follows the official ProxyCLIP implementation. The background threshold for Pascal VOC21 ~\cite{pascalvoc}, Pascal Context60 ~\cite{pascalcontext}, COCO-Object ~\cite{coco} is assigned to 0.2, 0.15, 0.25, for each.

\subsection{SCLIP Setting}
In SCLIP ~\cite{sclip} setting in Tab ~\ref{tab:sota_ovs_comparison}, 
We resize input images with a short side of 336 and perform sliding-window inference with a 224 $\times$ 224 window and 112 stride. Only for the Cityscapes, we resize the short side of 560. We use rename trick on both background class and several other classes, following official SCLIP code.  The background threshold for Pascal VOC21 ~\cite{pascalvoc}, Pascal Context60 ~\cite{pascalcontext}, COCO-Object ~\cite{coco} is assigned to 0.1, 0.1, 0.1, for each.

\begin{figure}[t]
    \centering
    \vspace{-0.2cm}
    \includegraphics[width=\columnwidth]{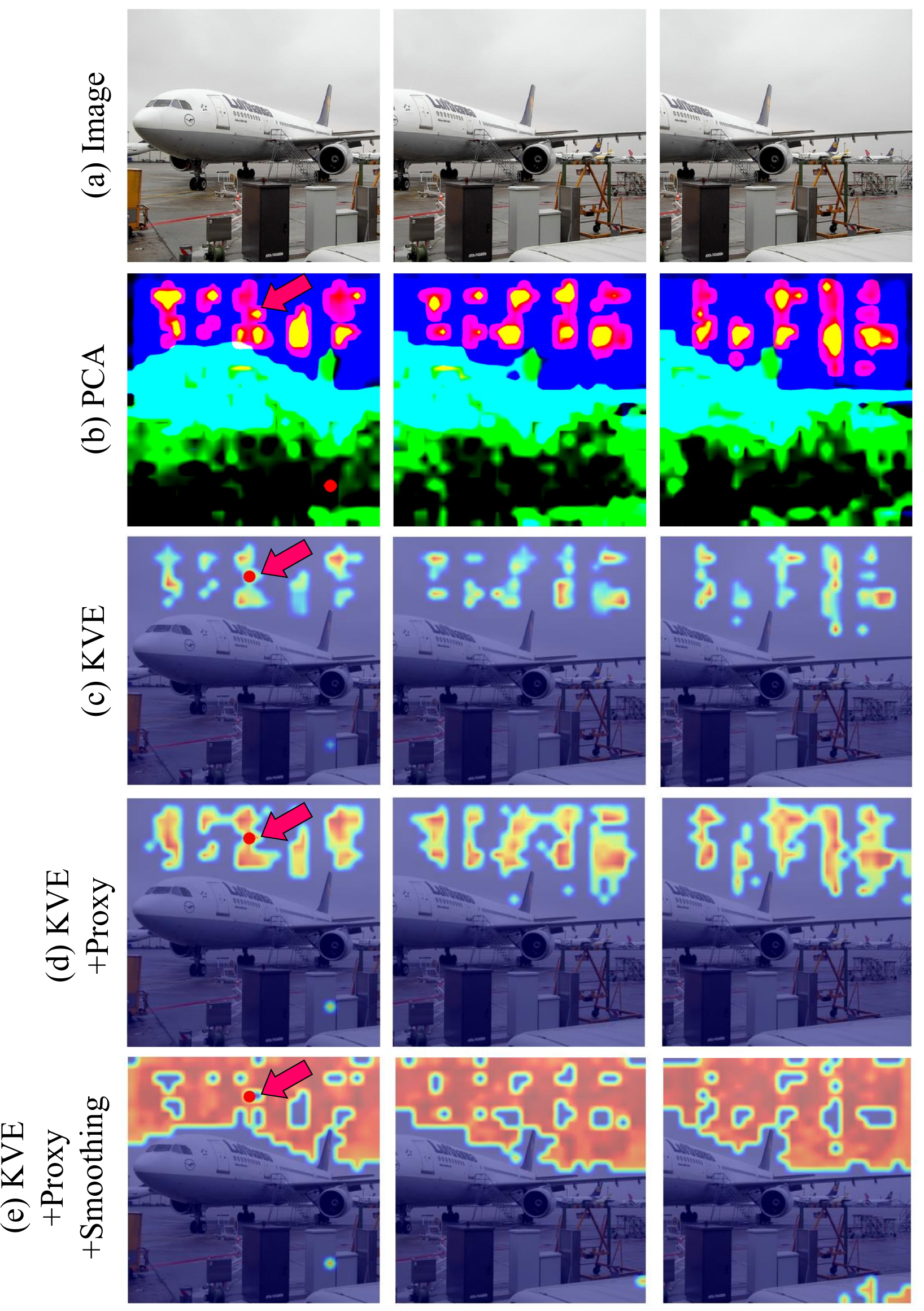}
    \vspace{-0.6cm}
    \caption{
        Effect of Smoothing on ClearCLIP attention map
    }
    \label{fig:ClearCLIP_Smoothing}
    \vspace{-0.5cm}
\end{figure}

\section{Query Smoothing for Proxy-Anchor Construction in CLIP features}
\label{sec:query_smoothing}

As we mentioned in the Implementation Details (Sec.~\ref{sec:experimental_settings}), our method incorporates a query smoothing step during proxy-anchor construction to ensure stable attention when generating attention maps from CLIP internal features, as in ClearCLIP~\cite{clearclip}. 
This query smoothing is necessary because CLIP feature maps contain noisy high-norm patches and these high-norm patches lost its semantic ~\cite{register}. By this reason high-norm patches do not have to be included in attention process. However, when the query token itself corresponds to a high-norm patch, the model tends to propagate this noise by repeatedly attending to other high-norm patches across windows, ultimately leading to incorrect semantic decisions. 
To mitigate this issue, we replace each raw query token with a smoothed token obtained by averaging it with neighbor patches. Specifically, we average each token with its eight spatial neighbors within the inner-window and with the patches at the same spatial position in adjacent overlapping windows:
\begin{align}
\mathbf{Q}_i^{(0)}
= \frac{
\mathbf{Q}_i
+ \sum_{j \in N(i)} \mathbf{Q}_j
+ \sum_{k \in O(i)} \mathbf{Q}_k
}{
1 + |N(i)| + |O(i)|
}.
\label{eq:smoothing}
\end{align}

Here, $N(i)$ denotes the local neighbors within the innner-window, and $O(i)$ represents the overlapping patches from adjacent outer-windows. This averaging suppresses high-norm activations of query token before the progressive positive mining step, allowing the model to aggregate meaningful positives across windows without reinforcing noisy high-norm responses.

Qualitatively, Fig.~\ref{fig:ClearCLIP_Smoothing} illustrates this process. 
In Fig.~\ref{fig:ClearCLIP_Smoothing}(a-b), the cropped images and its pca visualization identify the location of the high-norm patches. When adapting Key-Value Extension, high-norm patches in background regions (e.g., sky) activate attention both within inner and outer windows (Fig.~\ref{fig:ClearCLIP_Smoothing}(c)), and even with proxy-anchor stabilization, these noisy patches can dominate the attention map (Fig.~\ref{fig:ClearCLIP_Smoothing}(d)). By replacing query tokens with their smoothed versions (Eq.~\ref{eq:smoothing}), attention is redirected toward semantically meaningful tokens (Fig.~\ref{fig:ClearCLIP_Smoothing}(e)), demonstrating that our method effectively suppresses noisy high-norm patches and constructs stable attention maps.

\section{Semantic Collapse in Proxy Anchors}
We examine the potential risk of semantic collapse during proxy-anchor construction. In this process, a small number of negative tokens may be mistakenly selected as high-confident tokens, potentially leading to semantic collapse of the proxy anchor. To assess the robustness of our high-confident token filtering, we perform a binary classification using GT labels. As shown in Tab.~\ref{tab:high_confident_token_classification}, the lowest precision is 93.9\% on ADE20K~\cite{ade20k}, while all other datasets achieve even higher precision. These results indicate that semantic collapse rarely occurs, with at most 6\% of negative tokens being incorrectly incorporated into the proxy anchor.

\begin{table}[H]
\vspace{-0.2cm}
\centering
\scriptsize
\setlength{\tabcolsep}{3pt}
\renewcommand{\arraystretch}{1.1}

\begin{tabular}{
|>{\centering\arraybackslash}m{1.8cm}|
>{\centering\arraybackslash}m{0.7cm}|
>{\centering\arraybackslash}m{0.7cm}|
>{\centering\arraybackslash}m{0.7cm}|
>{\centering\arraybackslash}m{0.7cm}|
>{\centering\arraybackslash}m{0.7cm}|
>{\centering\arraybackslash}m{0.7cm}|}
\hline
Metric & V21 & PC59 & C-Stf & City & ADE & Avg. \\
\hline
\hline
Precision (\%) & 98.7 & 96.3 & 94.2 & 95.6 & \textbf{93.9} & 96.2 \\
\hline
\end{tabular}

\vspace{-0.2cm}
\caption{High confidence token classification precision.}
\label{tab:high_confident_token_classification}
\vspace{-0.4cm}
\end{table}

\section{Analysis for Attention Masking and Scaling}
\begin{figure}[t]
    \centering
    \vspace{-0.2cm}
    \includegraphics[width=\columnwidth]{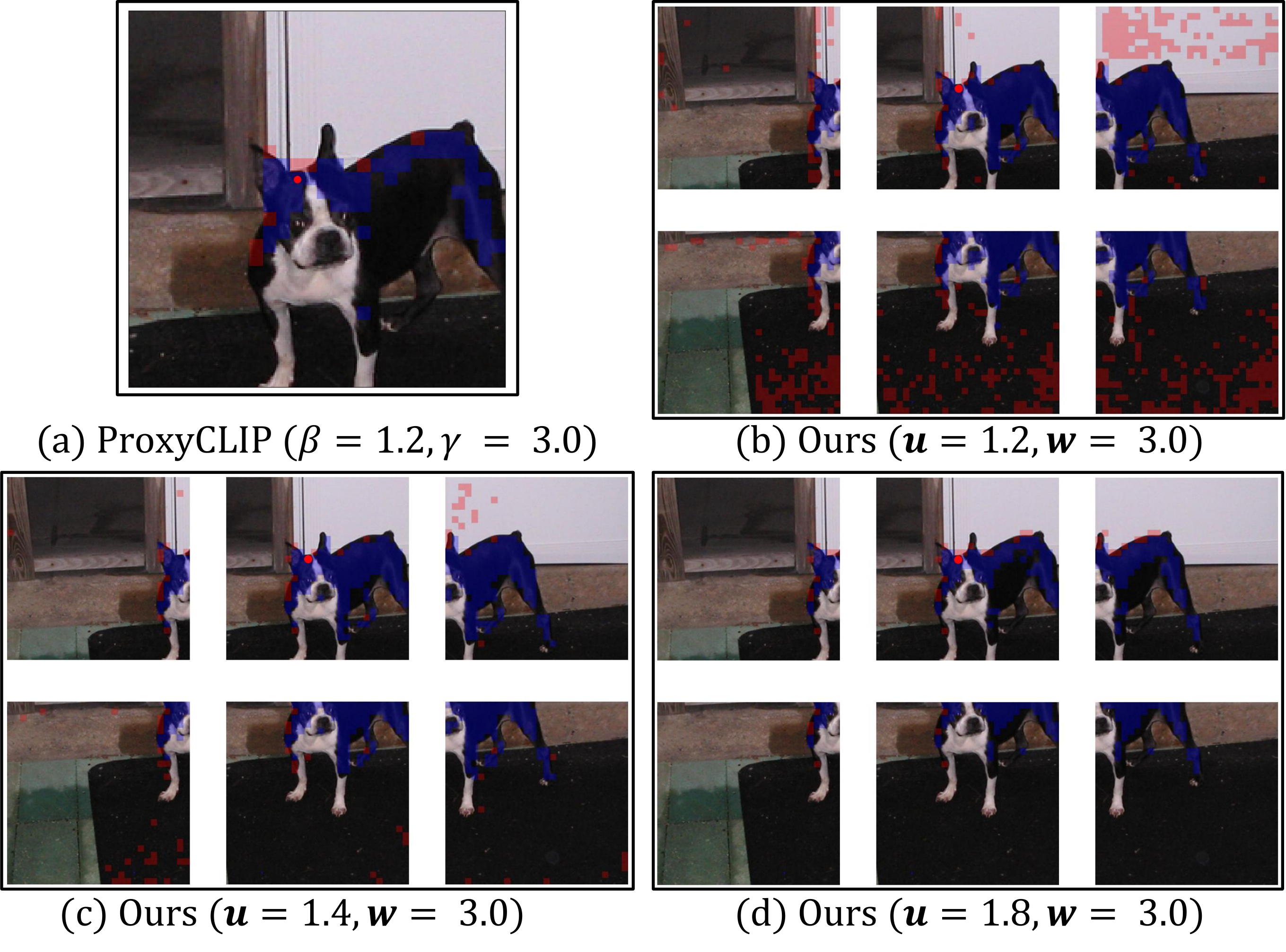}
    \vspace{-0.6cm}
    \caption{
        Effect of Key-Value Extension and \textbf{u}.
    }
    \label{fig:Effect_of_KVE_u}
    \vspace{-0.4cm}
\end{figure}

As discussed in Sec.~\ref{sec:dynamic_norm}, the Key–Value Extension can lead to the neglect of small objects due to the additional negative tokens from outer windows. Quantitatively, the total number of tokens increases by a factor of the number of windows \(L\), substantially altering the statistics of the similarity map. For small objects, most of the newly introduced tokens correspond to background (negative) regions. This circumstance is clearly shown in Fig.~\ref{fig:Effect_of_KVE_u}(a-b), where background tokens (red) remain largely unmasked in Fig.~\ref{fig:Effect_of_KVE_u}(b) compared with Fig.~\ref{fig:Effect_of_KVE_u}(a), illustrating that low values of \(\beta=1.2\) and \(\gamma=3.0\) fail to suppress negative tokens from the outer windows.

While increasing fixed \(\mathbf{u}\) and \(\mathbf{w}\) (which correspond to \(\beta\) and \(\gamma\) in ProxyCLIP~\cite{lan2024proxyclip}) can quantitatively suppress this additional negative tokens, setting them too high can over-mask the attention. This over-masking removes parts of the object itself, restricting attention to local sub-regions and ultimately degrading performance (see Fig.~\ref{fig:Effect_of_KVE_u}c-d).
 Conversely, moderately stronger settings such as \((\mathbf{u}=1.4, \mathbf{w}=5.0)\) achieve a balance: they suppress the negative tokens introduced by Key–Value Extension while preserving full object coverage, as evidenced qualitatively in Fig.~\ref{fig:Effect_of_KVE_u}.

These results directly support the normalization tightening strategy discussed in Sec.~\ref{sec:dynamic_norm} of the main paper. Specifically, the “best” results reported in Tab.~\ref{tab:sota_ovs_comparison} and Tab.~\ref{tab:ablation_study} are obtained by tightening the normalization parameters in this manner, compensating for the substantial increase of negative tokens introduced by Key–Value Extension. The observations from this results highlight a fundamental limitation of fixed masking and scaling: no single set of static hyperparameters can robustly handle both small and large objects across varying window configurations. This motivates the introduction of Dynamic Normalization, which adaptively adjusts masking and scaling to maintain consistent attention behavior regardless of object scale, window count, or dataset characteristics. Lastly, Dynamic Normalization removes the need for dataset-specific hyperparameters, providing a single formulation that generalizes well across datasets and window configurations.

% \begin{table}[t]
% \vspace{-0.2cm}
% \centering
% \scriptsize
% \setlength{\tabcolsep}{2pt}
% \renewcommand{\arraystretch}{1.2}

% \begin{tabular}{
% |>{\centering\arraybackslash}m{1.9cm}|
% >{\centering\arraybackslash}m{0.60cm}|
% >{\centering\arraybackslash}m{0.60cm}|
% >{\centering\arraybackslash}m{0.60cm}|
% >{\centering\arraybackslash}m{0.60cm}|
% >{\centering\arraybackslash}m{0.60cm}|
% >{\centering\arraybackslash}m{0.60cm}|
% >{\centering\arraybackslash}m{0.60cm}|
% >{\centering\arraybackslash}m{0.60cm}|
% }
% \hline
% \makecell{Model} &
% \makecell{V21} &
% \makecell{PC\\60} &
% \makecell{C-\\Obj} &
% \makecell{V20} &
% \makecell{PC\\59} &
% \makecell{C-\\Stf} &
% \makecell{City} &
% \makecell{ADE} \\
% \hline

% \makecell{ProxyCLIP}
% & 0.240 & 0.174 & 0.173 & 0.240 & 0.174 & 0.173 & 0.256 & 0.228 \\
% \hline

% \makecell{ProxyCLIP + KVE}
% & \textbf{0.214} & \textbf{0.141} & \textbf{0.142} & \textbf{0.214} & \textbf{0.143} & \textbf{0.142} & \textbf{0.208} & \textbf{0.205} \\
% \hline

% \end{tabular}

% \vspace{-0.2cm}
% \caption{Effect of Key–Value Extension on similarity statistics.}
% \label{tab:kv_extension_stats}
% \vspace{-0.1cm}
% \end{table}

\section{High confidence tokens}
\label{sec:high_confidence_tokens}
% High confidence token count isn't universally accurate proxy for object type; our dynamic norm is effective heuristic. Beyond class-wise analysis in Section~\ref{sec:object_scale}, we group instances by GT object size and show corresponding high confidence token counts (Fig.~\ref{fig:Object scale}), avoiding confounding factors like long-tailed class or texture-rich backgrounds and showing effectiveness as object-scale proxy.

\begin{figure}[H]
    \centering
    \vspace{-0.2cm}
    \includegraphics[width=\columnwidth]{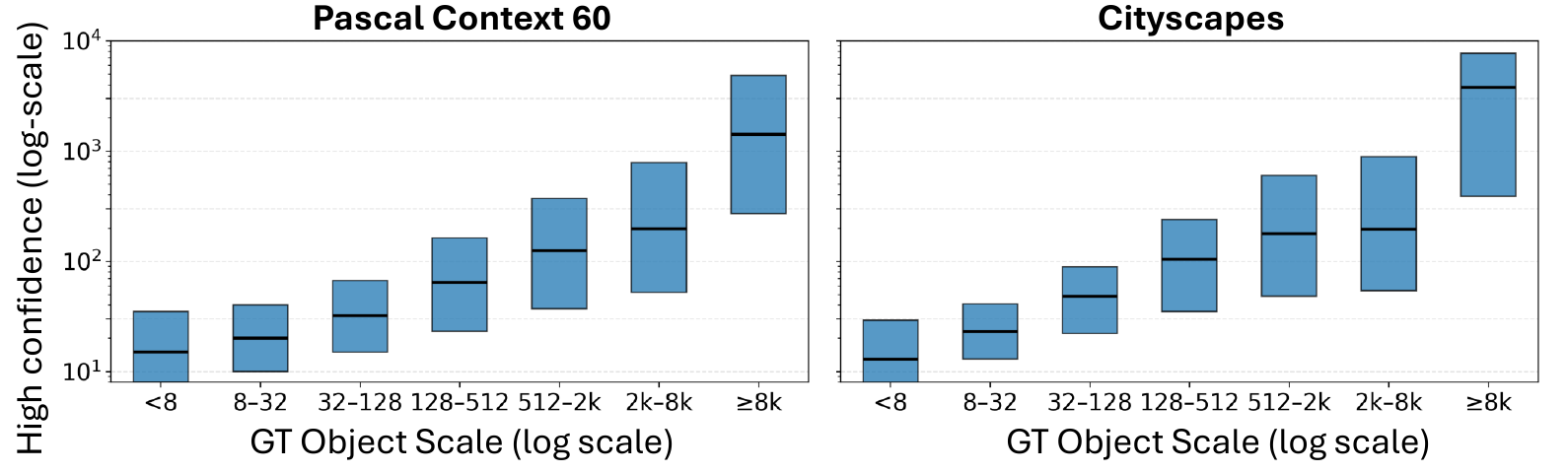}
    \vspace{-0.8cm}
    \caption{
        Object scale (GT vs high Confidence).
    }
    \label{fig:Object scale}
    \vspace{-0.35cm}
\end{figure}

High confidence token count is not a universally accurate proxy for object type; our dynamic normalization serves as an effective heuristic. Beyond the class-wise analysis in Section~\ref{sec:object_scale}, we group instances based on ground-truth (GT) object size and analyze the corresponding number of high confidence tokens in Pascal Context60~\cite{pascalcontext} and Cityscapes~\cite{cordts2016cityscapes} (in Fig.~\ref{fig:Object scale}). Specifically, for each image, we compute the number of high confidence tokens associated with each GT instance, assign it to a size bin according to its GT region size, and report the interquartile range (25th–75th percentile) and median within each bin. This analysis shows that the number of high confidence tokens provides a reliable proxy for object size, demonstrating its effectiveness in capturing scale information.

\vspace{-0.2cm}

\section{Adapting to SAM-based models}
We conduct experiments to evaluate the adaptability of our method to SAM-based OVSS model ~\cite{trident}. Our method can be integrated into SAM-based models, which utilize components such as SAM masks and SAM refinement. While SAM-based models generally exhibit lower inference speed, they achieve superior performance. We aim to demonstrate that our method is compatible with SAM-based models and has the potential to achieve state-of-the-art performance in this setting. As shown in Tab.~\ref{tab:Trident_GLA}, our method attains 46.1\% mIoU, representing a +0.3\% improvement over the baseline, Trident, which is SAM-based model. This result confirms that our method can be effectively adapted to SAM-based models, and that our contribution is independent of existing approaches.

\begin{table}[H]
\vspace{-0.2cm}
\centering
\scriptsize
\setlength{\tabcolsep}{2pt}
\renewcommand{\arraystretch}{1.1}

\begin{tabular}{
|>{\centering\arraybackslash}m{1.9cm}|
>{\centering\arraybackslash}m{0.52cm}|
>{\centering\arraybackslash}m{0.52cm}|
>{\centering\arraybackslash}m{0.52cm}|
>{\centering\arraybackslash}m{0.52cm}|
>{\centering\arraybackslash}m{0.52cm}|
>{\centering\arraybackslash}m{0.52cm}|
>{\centering\arraybackslash}m{0.52cm}|
>{\centering\arraybackslash}m{0.52cm}|
>{\centering\arraybackslash}m{0.52cm}|}
\hline
\makecell{Model} &
\makecell{V21} &
\makecell{PC\\60} &
\makecell{C-\\Obj} &
\makecell{V20} &
\makecell{PC\\59} &
\makecell{C-\\Stf} &
\makecell{City} &
\makecell{ADE} &
\makecell{Avg} \\
\hline
\hline
Trident
& 67.1 & 38.6 & 41.1 & 84.5 & 42.2 & \textbf{28.3} & 42.9 & \textbf{21.9} & 45.8 \\
\hline
\cellcolor{gray!15}\textbf{Trident + GLA}
& \cellcolor{gray!15}\textbf{67.4}
& \cellcolor{gray!15}\textbf{38.9}
& \cellcolor{gray!15}\textbf{41.4}
& \cellcolor{gray!15}\textbf{84.8}
& \cellcolor{gray!15}\textbf{42.4}
& \cellcolor{gray!15}\textbf{28.3}
& \cellcolor{gray!15}\textbf{43.6}
& \cellcolor{gray!15}21.7
& \cellcolor{gray!15}\textbf{46.1} \\
\hline
\end{tabular}

\vspace{-0.3cm}
\caption{Trident adaptation with GLA.}
\label{tab:Trident_GLA}
\vspace{-0.3cm}
\end{table}

\section{Adapting to Multi-layer Feature Fusion}
\label{sec:multi-layer_feature_fusion}
As we mentioned on Sec ~\ref{sec:generalization_across_backbones}, we conduct experiments to evaluate the compatibility of our method with multi-layer feature fusion in the training-free OVSS task, which has been widely adopted recently. We also examine whether our model demonstrates performance improvements when combined with multi-layer feature fusion, which can validate that our method independently enhances performance and can be considered a standalone technique. In Tab.~\ref{tab:multilayer_feature_fusion}, we compare our method with and without multi-layer feature fusion. Following the approach of SC-CLIP ~\cite{sc-clip}, which leverages multi-layer feature fusion, we observe that our method combined with multi-layer feature fusion achieves an average mIoU of 44.9\%, representing a +0.9\% improvement over our baseline. This result demonstrates that our method independently improves baseline performance and can be effectively adapted to multi-layer feature fusion.

\begin{table}[H]
\vspace{-0.2cm}
\centering
\scriptsize
\setlength{\tabcolsep}{2pt}
\renewcommand{\arraystretch}{1.1}

\begin{tabular}{
|>{\centering\arraybackslash}m{1.9cm}|
>{\centering\arraybackslash}m{0.52cm}|
>{\centering\arraybackslash}m{0.52cm}|
>{\centering\arraybackslash}m{0.52cm}|
>{\centering\arraybackslash}m{0.52cm}|
>{\centering\arraybackslash}m{0.52cm}|
>{\centering\arraybackslash}m{0.52cm}|
>{\centering\arraybackslash}m{0.52cm}|
>{\centering\arraybackslash}m{0.52cm}|
>{\centering\arraybackslash}m{0.52cm}|}
\hline
\makecell{Model} &
\makecell{V21} &
\makecell{PC\\60} &
\makecell{C-\\Obj} &
\makecell{V20} &
\makecell{PC\\59} &
\makecell{C-\\Stf} &
\makecell{City} &
\makecell{ADE} &
\makecell{Avg} \\
\hline
\hline
Ours
& 66.3 & 36.1 & 37.7 & \textbf{84.2} & 39.9 & 26.9 & 40.8 & 20.0 & 44.0 \\
\hline
\cellcolor{gray!15}\makecell{\textbf{Ours + Fusion}}
& \cellcolor{gray!15}\textbf{67.9}
& \cellcolor{gray!15}\textbf{37.0}
& \cellcolor{gray!15}\textbf{38.1}
& \cellcolor{gray!15}84.1
& \cellcolor{gray!15}\textbf{40.9}
& \cellcolor{gray!15}\textbf{27.5}
& \cellcolor{gray!15}\textbf{43.2}
& \cellcolor{gray!15}\textbf{20.3}
& \cellcolor{gray!15}\textbf{44.9} \\
\hline
\end{tabular}

\vspace{-0.3cm}
\caption{Effect of multi-layer feature fusion.}
\label{tab:multilayer_feature_fusion}
\vspace{-0.3cm}
\end{table}

\section{Adapting to CASS for SOTA Performance}
\label{sec:cass_adaptation}

As mentioned in Sec.~\ref{sec:generalization_across_backbones}, we evaluate the compatibility of our method with CASS~\cite{CASS} to demonstrate that text prompt embedding modifications can be effectively integrated. We adapt our method to CASS by combining GLA-CLIP with two key components of CASS: (1) Object-Guided Text Embedding Adjustment, which refines text embeddings by fusing them with the mean of visual tokens, and (2) Object Perspective Patch-Text Similarity, which processes the entire image instead of individual windows and incorporates the full-image logits into the final prediction. Using the dataset-specific hyperparameters provided in the official CASS code, our approach achieves state-of-the-art performance, with an average mIoU of 44.7\% in Tab.~\ref{tab:CASS_Adaptation}.
One of our contribution is that ours can be combined with text embedding modification and logit fusion.

\begin{table}[H]
\vspace{-0.2cm}
\centering
\scriptsize
\setlength{\tabcolsep}{2pt}
\renewcommand{\arraystretch}{1.2}

\begin{tabular}{
|>{\centering\arraybackslash}m{2.0cm}|
>{\centering\arraybackslash}m{0.53cm}|
>{\centering\arraybackslash}m{0.53cm}|
>{\centering\arraybackslash}m{0.53cm}|
>{\centering\arraybackslash}m{0.53cm}|
>{\centering\arraybackslash}m{0.53cm}|
>{\centering\arraybackslash}m{0.53cm}|
>{\centering\arraybackslash}m{0.53cm}|
>{\centering\arraybackslash}m{0.53cm}|
>{\centering\arraybackslash}m{0.53cm}|
}
\hline
\makecell{Model} &
\makecell{V21} &
\makecell{PC\\60} &
\makecell{C-\\Obj} &
\makecell{V20} &
\makecell{PC\\59} &
\makecell{C-\\Stf} &
\makecell{City} &
\makecell{ADE} &
\makecell{Avg} \\
\hline
\hline

CASS 
& 65.8 & 36.7 & \textbf{37.8} & 87.8 & 40.2 & 26.7 & 39.4 & 20.4  & 44.4 \\
ProxyCLIP + GLA 
& 66.3 & 36.1 & 37.7 & 84.2 & \textbf{40.8} & 26.9 & 39.9 & 20.0 & 44.0 \\
\hline
\cellcolor{gray!15}\makecell{\textbf{CASS + GLA}}
& \cellcolor{gray!15}\textbf{67.2}
& \cellcolor{gray!15}\textbf{36.8}
& \cellcolor{gray!15}36.2
& \cellcolor{gray!15}\textbf{88.5}
& \cellcolor{gray!15}39.7
& \cellcolor{gray!15}\textbf{27.2}
& \cellcolor{gray!15}\textbf{40.6}
& \cellcolor{gray!15}\textbf{21.0}
& \cellcolor{gray!15}\textbf{44.7} \\

\hline
\end{tabular}

\vspace{-0.25cm}
\caption{CASS adaptation with GLA}
\label{tab:CASS_Adaptation}
\vspace{-0.3cm}
\end{table}

% \begin{table}[t]
% \centering
% \scriptsize
% \begin{tabular}{l | c | c c c | c}
% \hline
% Model & Web (8) & Vaih. & UAVid & VDD & Avg (3) \\
% \hline
% \textbf{Ours} & \textbf{41.9} & 31.9 & 41.9 & 45.0 & \textbf{39.6} \\
% CLIP-DINOiser & \textbf{40.3} & 18.6 & 33.3 & 37.7 & \textbf{29.9} \\
% \hline
% \end{tabular}
% \vspace{-0.1cm}
% \caption{Domain-shift evaluation on remote sensing datasets.}
% \vspace{-0.2cm}
% \label{tab:domain_shift_results}
% \end{table}

\section{Domain Shift: Remote Sensing Datset}

\subsection{Evaluation on remote sensing dataset}
% To demonstrate that our GLA module is not only effective on the eight web datasets reported in Tab.~\ref{tab:sota_ovs_comparison}, but also generalizable to other domain, we evaluate ours on 3 remote sensing datasets, following the protocol of SegEarth-OV~\cite{segearth}. Specifically, since SegEarth-OV also addresses the OVSS task and employs a sliding-window approach, our module can be directly applied. Accordingly, we conduct OVSS on the Vaihingen, UAVid~\cite{uavid}, and VDD~\cite{vdd} datasets. As shown in Tab.~\ref{tab:segearth_ov}, incorporating the GLA module increases the average mIoU from 38.8\% to 39.6\%, corresponding to a +0.8\% improvement, demonstrating the module’s generality across domains. We expect that our module can enhance performance on any dataset that employs a sliding-window.

To demonstrate the generality of our GLA module beyond the eight web datasets in Tab.~\ref{tab:sota_ovs_comparison}, we evaluate it on three remote sensing datasets following SegEarth-OV~\cite{segearth}. Since SegEarth-OV also adopts a sliding-window OVSS pipeline, our module can be directly integrated. We conduct experiments on Vaihingen\footnote{https://www.isprs.org/education/benchmarks/UrbanSemLab}, UAVid~\cite{uavid}, and VDD~\cite{vdd}. As shown in Tab.~\ref{tab:segearth_ov}, GLA improves the average mIoU from 38.8\% to 39.6\% (+0.8\%), demonstrating strong cross-domain generalization. We expect our module to benefit any sliding-window-based setting.

\vspace{-0.2cm}

\begin{table}[H]
\centering
\scriptsize
\setlength{\tabcolsep}{2pt}
\renewcommand{\arraystretch}{1.15}

\begin{tabular}{
|>{\centering\arraybackslash}m{2.4cm}|
 >{\centering\arraybackslash}m{1.3cm}|
 >{\centering\arraybackslash}m{1.3cm}|
 >{\centering\arraybackslash}m{1.3cm}|
 >{\centering\arraybackslash}m{1.3cm}|}
\hline
\makecell{Model} &
\makecell{Vaih.} &
\makecell{UAVid} &
\makecell{VDD} &
\makecell{Avg} \\
\hline
\hline

ClearCLIP 
& 27.3 & 36.2 & 39.3 & 34.3 \\
ProxyCLIP
& 30.6 & 41.4 & 44.3 & 38.8 \\
\hline
\cellcolor{gray!15}\makecell{\textbf{ProxyCLIP + GLA}}
& \cellcolor{gray!15}\textbf{31.9}\,(\textbf{+1.3}) 
& \cellcolor{gray!15}\textbf{41.9}\,(\textbf{+0.5})
& \cellcolor{gray!15}\textbf{45.0}\,(\textbf{+0.7})
& \cellcolor{gray!15}\textbf{39.6}\,(\textbf{+0.8}) \\
\hline
\end{tabular}
\vspace{-0.2cm}
\caption{Performance comparison across remote sensing datasets.}
\vspace{-0.3cm}
\label{tab:segearth_ov}
\end{table}

\subsection{Comparison with training-based model}
Training-based OVSS models are vulnerable to domain shift due to reliance on fine-tuning. In domain-shift experiments (Tab.~\ref{tab:domain_shift_results}), ours and CLIP-DINOiser~\cite{clip-dinoiser} (which is the representative Training-based model) show small gap on 8 web-image datasets (e.g. pascal voc~\cite{pascalvoc}, ade20k~\cite{ade20k}, cityscapes~\cite{cordts2016cityscapes}, etc), it becomes larger on 3 remote sensing datasets, indicating limited robustness and motivating training-free OVSS that preserves zero-shot capability of CLIP. 
% Training-free methods incur no training cost, offering practical efficiency.
\vspace{-0.2cm}

\begin{table}[H]
\centering
\scriptsize
\setlength{\tabcolsep}{2pt}
\renewcommand{\arraystretch}{1.15}

\begin{tabular}{
|>{\centering\arraybackslash}m{2.0cm}|
 >{\centering\arraybackslash}m{1.2cm}|
 >{\centering\arraybackslash}m{1.1cm}
 >{\centering\arraybackslash}m{1.1cm}
 >{\centering\arraybackslash}m{1.1cm}|
 >{\centering\arraybackslash}m{1.1cm}|}
\hline
\makecell{Model} &
\makecell{Web (8)} &
\makecell{Vaih.} &
\makecell{UAVid} &
\makecell{VDD} &
\makecell{Avg (3)} \\
\hline
\hline

\cellcolor{gray!15}\makecell{\textbf{Ours}}
& \cellcolor{gray!15}\textbf{41.9} 
& \cellcolor{gray!15}{31.9}
& \cellcolor{gray!15}{41.9}
& \cellcolor{gray!15}{45.0}
& \cellcolor{gray!15}\textbf{39.6} \\
CLIP-DINOiser & \textbf{40.3} & 18.6 & 33.3 & 37.7 & \textbf{29.9} \\
\hline
\end{tabular}
\vspace{-0.2cm}
\caption{Domain-shift evaluation on remote sensing datasets.}
\vspace{-0.4cm}
\label{tab:domain_shift_results}
\end{table}

% 3. Latency & Memory Table
% \begin{table}[t]
% \centering
% \scriptsize
% \setlength{\tabcolsep}{2pt}  % column 간격 조정
% \renewcommand{\arraystretch}{0.9}  % row 높이 조정

% \begin{tabular}{|>{\centering\arraybackslash}p{0.45\columnwidth}|
%                 >{\centering\arraybackslash}p{0.25\columnwidth}|
%                 >{\centering\arraybackslash}p{0.25\columnwidth}|}
% \hline
% Latency (ms) / Memory (MB) & V21 & ADE \\
% \hline
% CASS & \textbf{542ms} / 2900MB & \textbf{2065ms} / 741MB  \\
% \hline
% ProxyCLIP & 136ms / 7560MB & 252ms / 8862MB \\
% ProxyCLIP + GLA & \textbf{138ms} / 9164MB & \textbf{258ms} / 9588MB \\
% \hline
% $\Delta$ (\%) & \textbf{+1\% }/ \textbf{+21\%} & \textbf{+2\%} / \textbf{+8\%} \\
% \hline
% \end{tabular}
% \vspace{-0.3cm}
% \caption{Latency and memory comparison.}
% \vspace{-0.5cm}
% \label{tab:latency_memory}
% \end{table}

\section{Computation cost.} 
\vspace{-0.1cm}
We analyze the computational cost of our model to assess its practical adaptability. 
Due to the Key--Value Extension, our method must forward batched images to reference tokens from outer-window; forwarding each window independently is not feasible. In addition, the Key--Value Extension increases the number of tokens involved in the attention computation. While ProxyCLIP~\cite{lan2024proxyclip} requires a complexity of $O(LN^{2}D)$, our approach incurs $O(L^{2}N^{2}D)$ complexity on attention operation.
Despite this theoretical increase, the actual latency and memory overhead remain modest.
We measure the number of sliding windows, latency, and memory usage on the Pascal Context59 ~\cite{pascalcontext} following the SCLIP~\cite{sclip} setting (Tab~\ref{tab:Latency_Memory_Context59}). We conduct the experiments on a NVIDIA RTX A6000 GPU. By varying the sliding-window stride, increasing the number of windows leads to higher memory usage in ours. Though even with more windows, ours remains more efficient than CASS~\cite{CASS}, using less memory while achieving competitive mIoU.
\vspace{-0.2cm}

\begin{table}[H]
\centering
\scriptsize
\setlength{\tabcolsep}{2pt}
\renewcommand{\arraystretch}{1.15}

\begin{tabular}{
|>{\centering\arraybackslash}m{2.7cm}|
 >{\centering\arraybackslash}m{0.9cm}|
 >{\centering\arraybackslash}m{1.1cm}
 >{\centering\arraybackslash}m{1.1cm}
 >{\centering\arraybackslash}m{1.1cm}|
}
\hline
\makecell{\textbf{Context59 (336px×497px)}} &
\makecell{\#Win} &
\makecell{Lat.} &
\makecell{Mem.} &
\makecell{mIoU} \\
\hline
\hline

CASS & 1 & 5799\,ms & 7.6\,GB & 40.2 \\
\hline
ProxyCLIP (Baseline) & 8 & 429\,ms & 1.8\,GB & 38.8 \\
\hline
Ours (Stride=224) & 6 & 415\,ms & 1.7\,GB & 39.5 \\
Ours (Stride=112) & 8 & 431\,ms & 1.9\,GB & 39.9 \\
Ours (Stride=98) & 12 & 441\,ms & 2.4\,GB & 39.9 \\
\hline

\end{tabular}
\vspace{-0.3cm}
\caption{Latency \& memory usage comparison on PC59}
\vspace{-0.3cm}
\label{tab:Latency_Memory_Context59}
\end{table}

\section{Setting ablation.}
\vspace{-0.1cm}
We tune model hyperparameters for each setting while applying the same configuration consistently to both the baseline and our method. 
As shown in Tab.~\ref{tab:voc21_inference_ablation}, our method consistently outperforms the ProxyCLIP~\cite{lan2024proxyclip} on Pascal VOC21~\cite{pascalvoc} across various configurations. 
The default setting is defined as \{crop size: 224, stride: 112, image size: 336, background threshold: 0.1, rename trick: enabled\} (following SCLIP~\cite{sclip} setting). 
We vary each parameter individually to demonstrate the robustness of our method.

% \begin{table}[H]
% \centering
% {\scriptsize
% \setlength{\tabcolsep}{2.5pt}
% \renewcommand{\arraystretch}{1.0}
% \begin{tabular}{l|c|ccccc}  % ← Default 열만 | |
% \hline
% \multirow{2}{*}{VOC21} 
% & \multirow{2}{*}{Default} 
% & Crop size 
% & Stride 
% & Img resize 
% & Bg. th. 
% & Rename \\
% & 
% & $224 \rightarrow 280$ 
% & $112 \rightarrow 224$
% & $336 \rightarrow 448$ 
% & $0.1 \rightarrow 0.2$ 
% & O$\rightarrow$X \\
% \hline
% ProxyCLIP & 63.3 & 64.4 & 59.7 & 61.8 & 61.0 & 60.4 \\
% Ours      & \textbf{66.3} & \textbf{66.5} & \textbf{63.6} & \textbf{65.3} & \textbf{64.6} & \textbf{63.3} \\
% \hline
% \end{tabular}}
% \vspace{-0.3cm}
% \caption{Setting ablation on VOC21 (mIoU).}
% \vspace{-0.4cm}
% \label{tab:voc21_inference_ablation}
% \end{table}

\begin{table}[H]
\vspace{-0.2cm}
\centering
\scriptsize
\setlength{\tabcolsep}{2pt}
\renewcommand{\arraystretch}{1.1}

\begin{tabular}{
|>{\centering\arraybackslash}m{1.1cm}|
>{\centering\arraybackslash}m{1.0cm}|
>{\centering\arraybackslash}m{1.0cm}|
>{\centering\arraybackslash}m{1.0cm}|
>{\centering\arraybackslash}m{1.2cm}|
>{\centering\arraybackslash}m{1.0cm}|
>{\centering\arraybackslash}m{1.0cm}| }
\hline
\multirow{3}{*}{VOC21} &
\multirow{3}{*}{Default} 
& Crop size 
& Stride 
& Img resize 
& Bg. th. 
& Rename \\
& 
& $224\rightarrow$ 
& $112\rightarrow$
& $336\rightarrow$ 
& $0.1\rightarrow$ 
& O$\rightarrow$ \\
&
& 280 
& 224
& 448 
& 0.2 
& X \\
\hline
\hline
ProxyCLIP & 63.3 & 64.4 & 59.7 & 61.8 & 61.0 & 60.4 \\
\hline
\cellcolor{gray!15}\makecell{\textbf{Ours}}
& \cellcolor{gray!15}\textbf{66.3}
& \cellcolor{gray!15}\textbf{66.5}
& \cellcolor{gray!15}\textbf{63.6}
& \cellcolor{gray!15}\textbf{65.3}
& \cellcolor{gray!15}\textbf{64.6}
& \cellcolor{gray!15}\textbf{63.3} \\
\hline
\end{tabular}

\vspace{-0.2cm}
\caption{Setting ablation on VOC21 (mIoU).}
\label{tab:voc21_inference_ablation}
\vspace{-0.2cm}
\end{table}

\vspace{-0.2cm}
\begin{figure*}[t]
    \centering
    \vspace{-0.2cm}
    \includegraphics[width=0.9\textwidth]{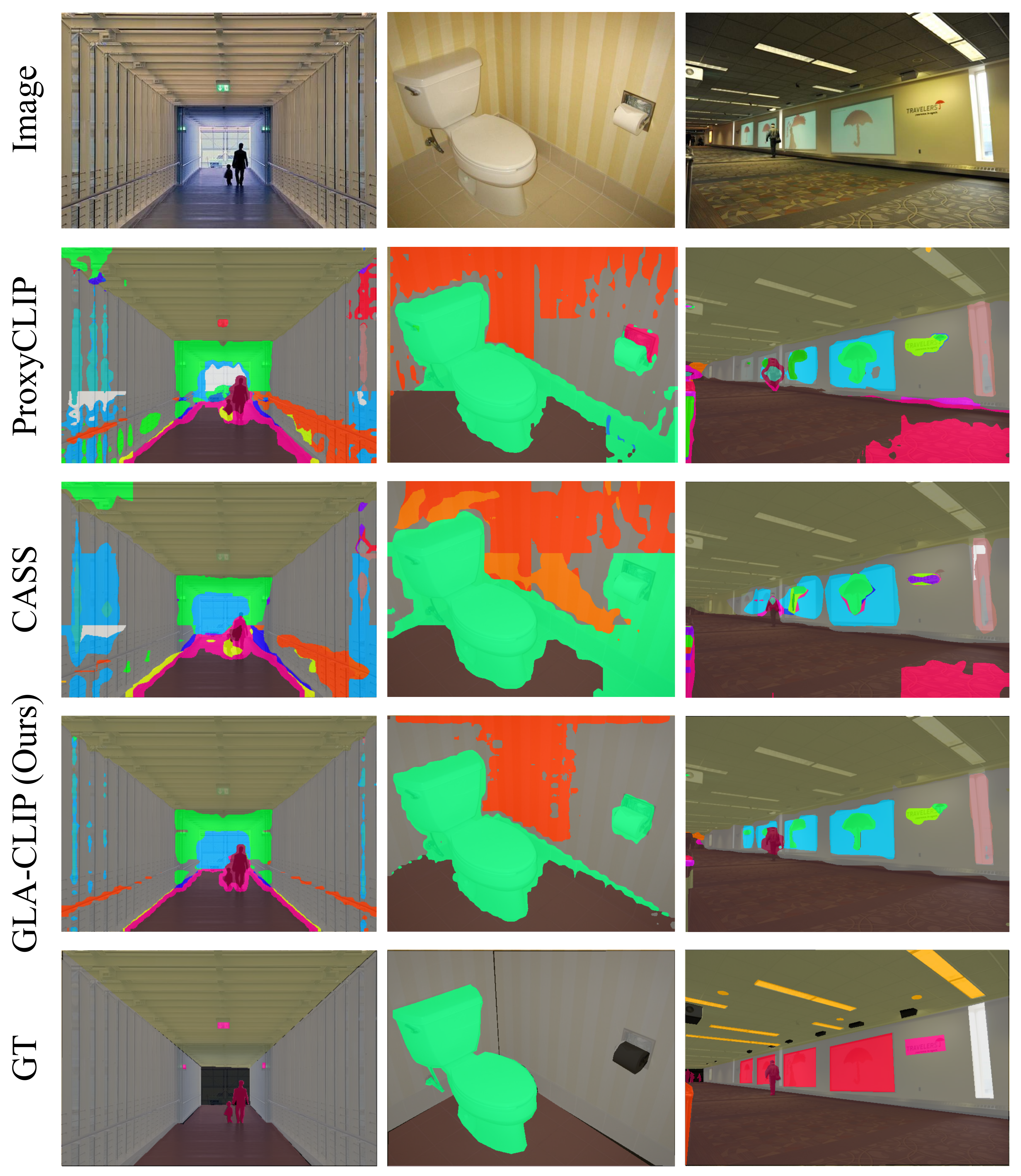}
    \vspace{-0.4cm}
    \caption{
        Qualitative results on ADE20K~\cite{ade20k} with 150 categories. We compare GLA-CLIP with ProxyCLIP~\cite{lan2024proxyclip}, CASS~\cite{CASS}.
    }
    \label{fig:ADE20k_vis}
    \vspace{-0.35cm}
\end{figure*}

\vspace{-0.2cm}
\begin{figure*}[t]
    \centering
    \vspace{-0.2cm}
    \includegraphics[width=0.9\textwidth]{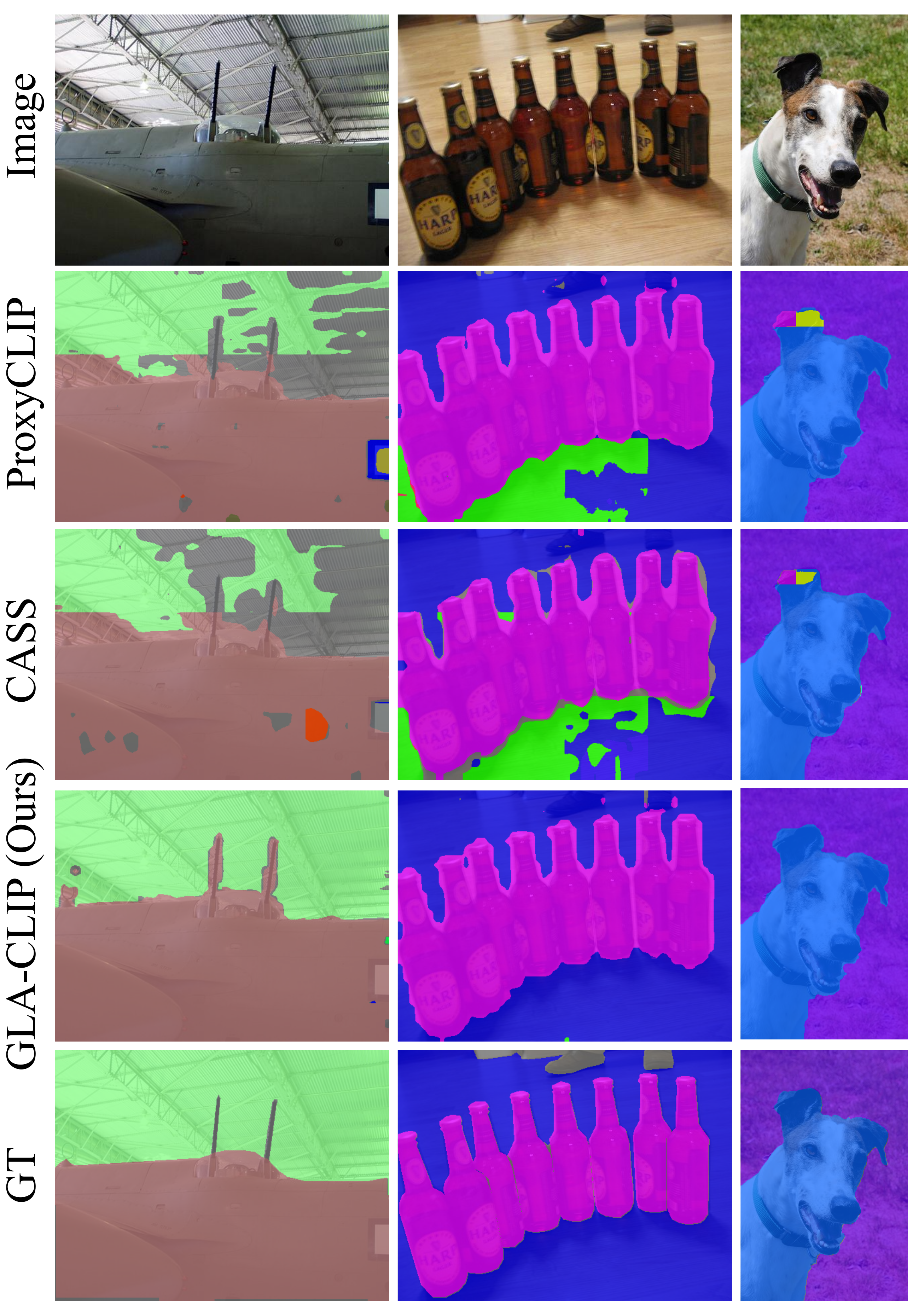}
    \vspace{-0.4cm}
    \caption{
        Qualitative results on Pascal Context60 ~\cite{pascalcontext} with 60 categories.. We compare GLA-CLIP with ProxyCLIP~\cite{lan2024proxyclip}, CASS~\cite{CASS}.
    }
    \label{fig:Context60_vis}
    \vspace{-0.35cm}
\end{figure*}

\vspace{-0.2cm}
\begin{figure*}[t]
    \centering
    \vspace{-0.2cm}
    \includegraphics[width=0.9\textwidth]{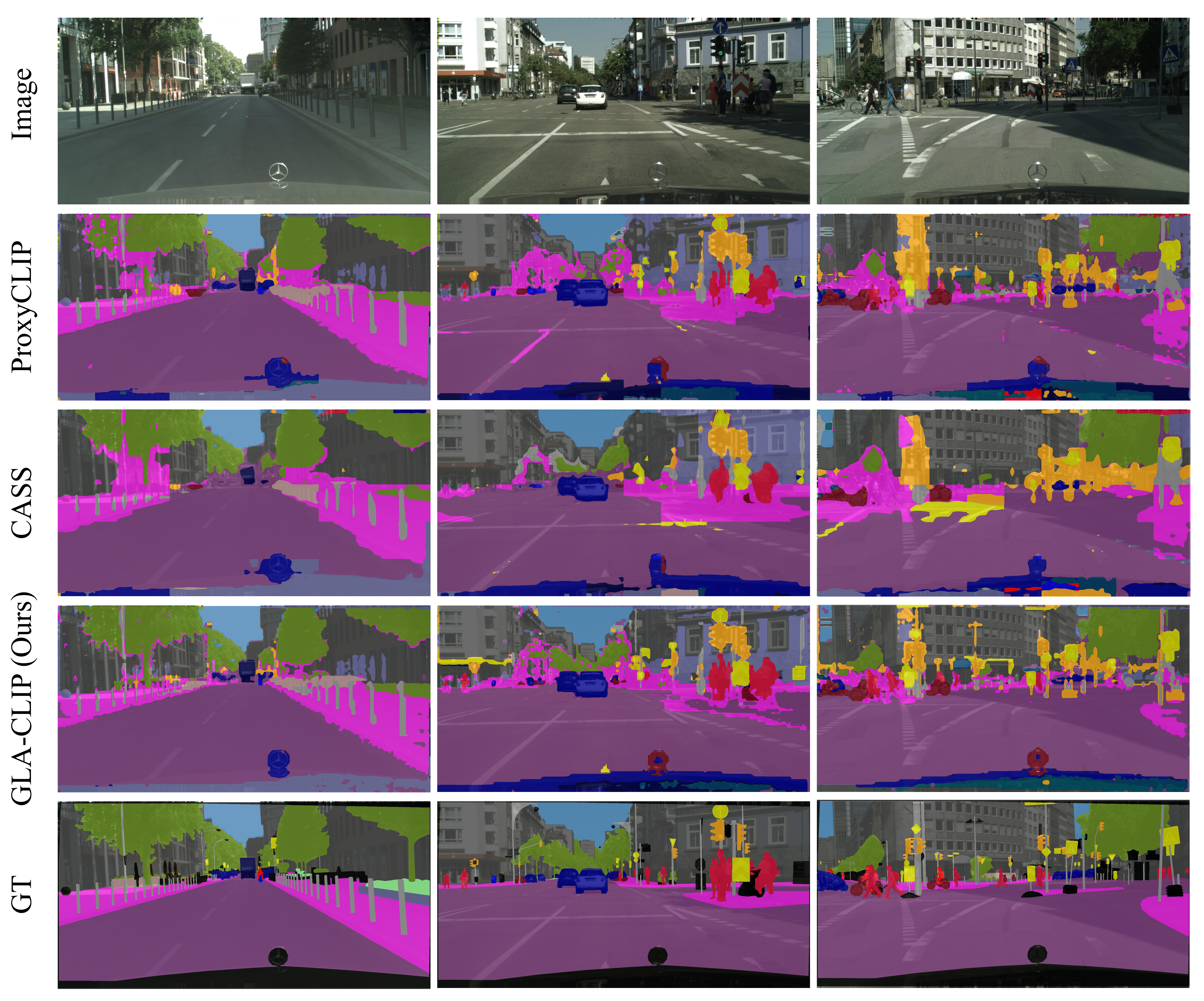}
    \vspace{-0.4cm}
    \caption{
        Qualitative results on Cityscapes~\cite{cordts2016cityscapes} with 19 categories. We compare GLA-CLIP with ProxyCLIP~\cite{lan2024proxyclip}, CASS~\cite{CASS}.
    }
    \label{fig:Cityscapes_vis}
    \vspace{-0.35cm}
\end{figure*}
% {
%     \clearpage
%     \small
%     \bibliographystyle{ieeenat_fullname}
%     \bibliography{supplement} % main.bib 사용

\end{document}